\documentclass{article}

\usepackage{microtype}
\usepackage{graphicx}
\usepackage{subfigure}
\usepackage{booktabs} 
\usepackage{multirow}
\usepackage{multicol}
\usepackage{colortbl}
\usepackage{xcolor}
\usepackage{float}
\usepackage{rotating}
\usepackage{placeins}
\usepackage{lscape}
\usepackage{adjustbox}
\usepackage{hyperref}

\usepackage[accepted]{icml2025}

\usepackage{amsmath}
\usepackage{amssymb}
\usepackage{mathtools}
\usepackage{amsthm}
\usepackage{bm}
\usepackage{pifont}

\newcommand{\E}{\mathbb{E}} 
\DeclareMathOperator*{\argmin}{arg\,min}

\usepackage[capitalize,noabbrev]{cleveref}

\theoremstyle{plain}

\theoremstyle{definition}

\theoremstyle{remark}

\usepackage[textsize=tiny]{todonotes}

\begin{document}

\twocolumn[
\icmltitle{Subgroups Matter for Robust Bias Mitigation}



\icmlsetsymbol{equal}{*}

\begin{icmlauthorlist}
\icmlauthor{Anissa Alloula}{ox}
\icmlauthor{Charles Jones}{ic}
\icmlauthor{Ben Glocker}{ic}
\icmlauthor{Bart\l omiej W. Papie\.z}{ox}

\end{icmlauthorlist}

\icmlaffiliation{ox}{University of Oxford, UK}
\icmlaffiliation{ic}{Imperial College London, UK}

\icmlcorrespondingauthor{Anissa Alloula}{anissa.alloula@dtc.ox.ac.uk}

\icmlkeywords{Machine Learning, ICML, Robustness, Fairness, Spurious Correlations, Bias Mitigation, Generalisation}

\vskip 0.3in
]



\printAffiliationsAndNotice{}  
 
\begin{abstract} 
Despite the constant development of new bias mitigation methods for machine learning, no method consistently succeeds, and a fundamental question remains unanswered: when and why do bias mitigation techniques fail? In this paper, we hypothesise that a key factor may be the often-overlooked but crucial step shared by many bias mitigation methods: the definition of subgroups. To investigate this, we conduct a comprehensive evaluation of state-of-the-art bias mitigation methods across multiple vision and language classification tasks, systematically varying subgroup definitions, including coarse, fine-grained, intersectional, and noisy subgroups. Our results reveal that subgroup choice significantly impacts performance, with certain groupings paradoxically leading to worse outcomes than no mitigation at all. Our findings suggest that observing a disparity between a set of subgroups is not a sufficient reason to use those subgroups for mitigation. Through theoretical analysis, we explain these phenomena and uncover a counter-intuitive insight that, in some cases, improving fairness with respect to a particular set of subgroups is best achieved by using a different set of subgroups for mitigation. Our work highlights the importance of careful subgroup definition in bias mitigation and presents it as an alternative lever for improving the robustness and fairness of machine learning models.

\end{abstract}

\section{Introduction}
\label{intro}
A significant barrier to the wider deployment of machine learning (ML) models is their tendency to fail when tested on distributions that differ from their training data. One particularly concerning manifestation of this issue is performance degradation for population subgroups, often caused by bias in training data such as spurious correlations (SC), under-representation of certain subgroups, or shifts in the presentation of the target $Y$ \cite{jonesCausalPerspectiveDataset2024a}. Bias mitigation methods aim to address these issues by training more robust models which are less susceptible to these biases, thereby improving generalisation. These methods generally adapt model training to improve the performance of some disadvantaged subgroups within the training data.

Despite the number of bias mitigation methods which have been proposed, benchmarks are increasingly reporting that the performance of these methods is inconsistent when tested in new settings, and that they often fail to surpass the empirical risk minimisation (ERM) baseline \cite{zongMEDFAIRBENCHMARKINGFAIRNESS2023,zietlowLevelingComputerVision2022,chenComprehensiveEmpiricalStudy2023,shresthaInvestigationCriticalIssues2022a,10.1007/978-3-031-72787-0_14}. Some efforts have been made to begin to elucidate the conditions under which certain bias mitigation methods might be valid, such as the work of \citet{jonesRethinkingFairRepresentation2024} in fair representation learning and \citet{schrouffMindGraphWhen2024} in data balancing. However, the choice of an appropriate mitigation method is only one aspect of the problem. To successfully mitigate bias, we must also select which subgroups we wish to apply the methods on -- a question which very little work has explicitly addressed.

Indeed, most bias mitigation methods rely on some form of grouping to first identify disadvantaged subgroups within the training data and then to implement group-based strategies aimed at improving generalisation or fairness. This can be as simple as observing a disparity in model performance between men and women, and trying to fix this by rebalancing the training data such that both subgroups are more uniformly distributed \cite{wengAreSexbasedPhysiological2023}, or noticing that a model performs poorly on data coming from a specific type of scanner, and applying a robust learning strategy such as adversarial training to prevent learning of scanner-specific but task-irrelevant information \cite{ganinunsuperviseddomainadapt2015}. To date, the literature has predominantly focused on simple and coarse subgroups, for example blonde hair and non-blonde hair in the CelebA dataset \cite{liuDeepLearningFace} or waterbirds and landbirds in the waterbird dataset \cite{sagawaDistributionallyRobustNeural2020}. In medical applications, the subgroups are often ``white" or ``non-white", ``men" or ``women", or some coarse binning of age \cite{riccilaraAddressingFairnessArtificial2022}. Subgroup choice is often motivated by two factors: a) practical constraints, for instance only having annotations for common attributes, and b) ethical or societal goals to achieve fairness with respect to specific subgroups. However these subgroups may poorly capture the underlying cause of model underperformance, thus obscuring critical information for bias mitigation methods.

In this work, we aim to better understand whether we can optimise this crucial step of subgroup definition in the same way that new bias mitigation methods are optimised. We investigate the role of subgroup definition on the performance of these methods, and  whether poor subgroup definition might explain why these methods often fail. We construct a setting of bias inspired by a real-world chest radiograph example \cite{olesenSlicingBiasExplaining2024} where there is a spurious correlation during training but which is absent during testing, and which is present in different proportions across subgroups, resulting in disparities in model performance. 
We consider realistic ways in which subgroups might be generated based on relevant attributes, and explore how they impact the performance of four state-of-the-art bias mitigation methods in four semi-synthetic vision and language datasets. 
We identify key patterns in the performance of these methods across groupings. Certain groupings lead to a large improvement over the ERM baseline, while others substantially lower performance relative to the baseline. We propose that the effectiveness of a  given subgrouping strategy is linked to its ability to recover the unbiased test distribution. We summarise the key contributions of this work as follows:
\begin{itemize}
    \item We show that the groupings used for bias mitigation strongly affect how well each method works, and provide insights on optimal grouping strategies.
    \item We argue that observing a disparity in model performance across a set of subgroups does not justify using those subgroups for mitigation, and may in fact, make matters worse.
    \item We provide a possible explanation for the differences in subgroup effectiveness based on the minimum KL divergence between the subgroup-weighted biased distribution and the unbiased test distribution.
    \item We challenge the conventional assumption that the best way to obtain ``fairness'' with respect to a specific set of subgroups is always achieved by using those same subgroups for bias mitigation. 
\end{itemize}

\section{Related work}
\label{related_work}
\subsection{Bias identification}
Research on bias detection has increasingly focused on refining subgroup definitions to capture complex patterns of unfairness. While individual fairness, as introduced by \citet{dworkFairnessAwareness2012}, offers a theoretically elegant approach by evaluating fairness at the individual level, its practical challenges have limited its adoption. As a result, group-based analyses remain the dominant paradigm. More recently, efforts have been made to move beyond traditional binary categories (e.g., men/women or white/non-white) to identify disparities that such coarse classifications may obscure.  For instance, \citet{kearnsPreventingFairnessGerrymandering2018} and \citet{buolamwiniGenderShadesIntersectional2018} illustrate how failing to account for intersections of attributes can entirely hide performance disparities. \citet{xuIntersectionalUnfairnessDiscovery2024} discuss the difficulty of identifying intersectional bias when there are a number of attributes at play, and propose a generative approach to discover high-bias intersections amongst many possibilities. 
Similarly, \citet{movvaCoarseRaceData2023} demonstrate that in clinical risk prediction models, variation in performance \textit{within} 4 commonly-used ethnic groups often exceeds the variation \textit{between} these coarse groups, advocating for the use of much more precise categories to describe ethnicity. These works highlight how heterogenous subgroups on which unfairness is observed can be, yet most of them do not consider how this translates to conducting bias mitigation.

    
\subsection{Subgroup definition for bias mitigation}
A smaller body of work has considered how this problem of subgroup definition extends to bias mitigation. For example, \citet{awasthiEqualizedOddsPostprocessing2020}, \citet{wangRobustOptimizationFairness2020a}, and \citet{stromberg2024for} explore how noise in subgroup annotations may impact post-processing, distributionally robust optimisation (DRO), and last-layer-retraining respectively. They find that fairness is not guaranteed under some perturbations of the true attributes. \citet{ghosalDistributionallyRobustOptimization2023} also extend group DRO (gDRO) with probabilistic subgroup labels if there is uncertainty with respect to the subgroup annotations. \citet{liWhacAMoleDilemmaShortcuts2023,kimImprovingRobustnessMultiple2024} show the difficulty of mitigating bias when there are multiple spurious correlations and annotated subgroups. Perhaps the work most closely related to ours is \citet{zhouExaminingCombatingSpurious2021}, which considers a setting where a spurious correlation is causing bias, and show that gDRO fails in an example where the subgroups do not directly account for the spurious correlation. They attribute this failure to the inability to upweight bias-conflicting samples effectively, as their constructed ``imperfect'' groups also include spuriously correlated samples - though their subgroup construction appears somewhat unrealistic. All of these works point to a sensitivity of bias mitigation methods to the subgroups defined, however they each only explore one possible flaw in subgroup annotations and restrict their scope to a single bias mitigation method.

\subsection{Bias mitigation without subgroups}
While the above works explore some possible failures in subgroup definition for common bias mitigation methods, others have developed new methods altogether which do not require subgroups to be defined in the traditional way. For instance, \citet{kearnsPreventingFairnessGerrymandering2018} and \citet{pmlr-v80-hebert-johnson18a} propose algorithms which aim to achieve fairness across all identifiable or richly structured subgroup classes. Bias discovery methods bypass the question of pre-defining subgroups altogether, and are useful in settings where subgroup annotations are missing. Some methods decouple bias discovery from bias mitigation by first ``discovering’’ biases through inferring subgroup annotations with an external model \cite{hanImprovingGroupRobustness2024,maraniViGBiasVisuallyGrounded2024} or by clustering points based on their feature representations \cite{krishnakumarUDISUnsupervisedDiscovery}, and subsequently performing bias mitigation. Other methods forgo explicit subgroup identification and instead rely on some form of regularisation or upweighting of misclassified samples \cite{ahnMitigatingDatasetBias2022,parkSelfsupervisedDebiasingUsing2022,liuJustTrainTwice2021}. 

We restrict our main analysis to established mitigation methods which use subgroup annotations because (a), they often serve as the upper bound for mitigation methods \cite{zhouExaminingCombatingSpurious2021,pezeshkiGradientStarvationLearning2021, bayasiBiasPrunerDebiasedContinual2024}, (b), many nominally label-free methods still require subgroup annotations for validation or hyperparameter tuning \cite{pezeshki2024discovering}, and (c), label-free methods frequently infer subgroups, making it still essential to understand their role.



\section{Background}
\label{background}
\subsection{Overview of bias mitigation methods}
\begin{figure*}[ht]
    \centering
{\includegraphics[width=0.9\textwidth]{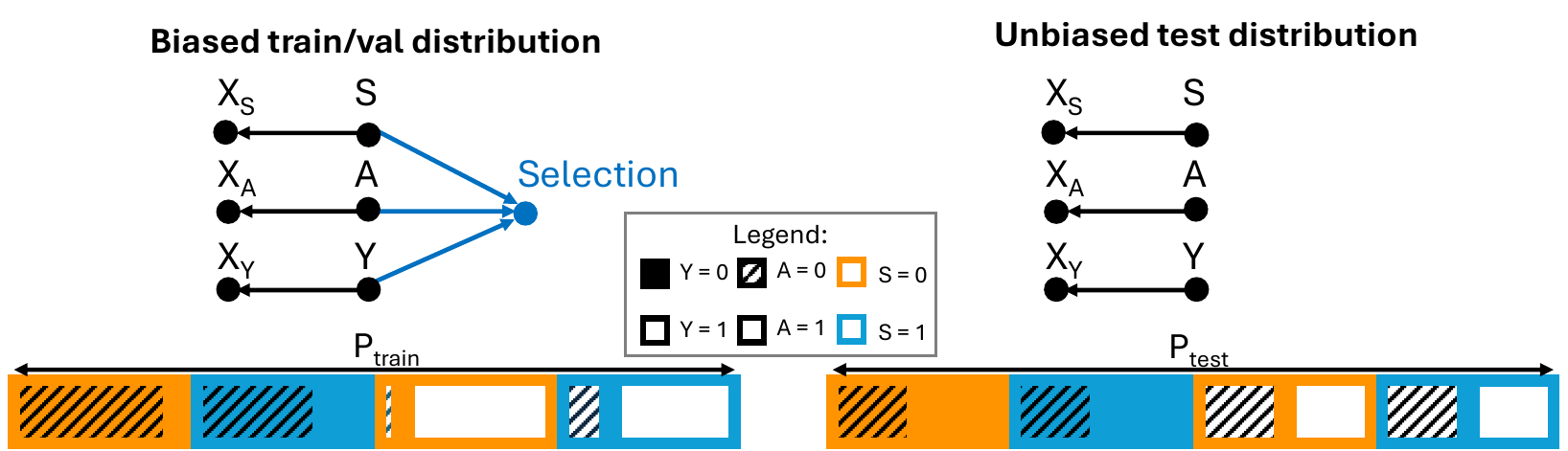}}
    \caption{Causal graphs representing interactions between $Y$, $A$, and $S$ variables in the training data and in the unbiased test data. Conditioning on selection in the training data results in spurious correlations between $Y$, $A$, and $S$. Coloured bars also illustrate the proportions of $Y$, $A$, and $S$ combinations in both settings.}
    \label{fig:causal graphs and dist diagrams}
\end{figure*}

\subsubsection{Empirical risk minimisation}

Traditional deep learning models conduct empirical risk minimisation (\textbf{ERM}), where, for given inputs $x \in \mathcal{X}$ and labels $y \in \mathcal{Y}$ from a distribution $\mathcal{P}$, the objective is to find a model $\theta \in \Theta$, that minimises the expected loss $\E_\mathcal{P}[\ell(\theta; (x, y))]$ . However, in practice, only a subset of $\mathcal{P}$, denoted $\mathcal{P}_{train}$, is available, so the loss over all samples in $\mathcal{P}_{train}$ is minimised:
\begin{align}\label{eq: erm}
  \hat{\theta}_{\text{ERM}} := \argmin_{\theta \in \Theta} \ \E_{(x, y) \sim {\mathcal{P}_{train}}}[\ell(\theta; (x, y))].
\end{align}

\subsubsection{Common bias mitigation methods}

Minimising the average loss over $\mathcal{P}_{train}$ often results in poor generalisation and poor performance on minority subgroups in the data. To address these shortcomings, bias mitigation methods have been proposed as alternatives to ERM. In this work, we focus on 4 commonly used bias mitigation methods which have demonstrated state-of-the-art results in certain tasks and represent the broad spectrum of existing methods. We define $k$ disjoint subgroups $\mathcal{P}_{g}$ indexed by $\mathcal{G} = \{1,...,k\}$, which partition
$\mathcal{P}_{train}$. We assume each training sample is annotated with its subgroup label $g$, thus giving the tuples $(x,y,g)$; however, subgroup information may be unavailable at inference time.

\textbf{Group distributionally robust optimisation (gDRO) }\cite{sagawaDistributionallyRobustNeural2020} reweights samples during loss calculation. The objective is to minimise the worst-case expected loss across each subgroup $\mathcal{P}_g$, (Equation~\eqref{eq: gdro}). In practice, an online optimisation algorithm is used to assign higher weight to high-loss subgroups in each batch. 

\begin{align}\label{eq: gdro}
  \hat{\theta}_{\text{gDRO}} := \underset{\theta \in \Theta}{\vphantom{\sup}\argmin} \Bigl\{ \underset{g \in \mathcal{G}}{\vphantom{\argmin}\max} \ \E_{(x, y) \sim \mathcal{P}_{train_g}}[\ell(\theta; (x, y))] \Bigr\}.
\end{align}
 
\textbf{Resampling} \cite{idrissiSimpleDataBalancing2022} also relies on reweighting but achieves it by adjusting the sampling probability of each subgroup $\mathcal{P}_g$ at the batch level so that each batch is balanced across subgroups. The loss for resampling is equivalent to:
\begin{align}\label{resampling equation}
\hat{\theta}_{\text{resampling}} := \argmin_{\theta \in \Theta} \ \sum_{g=0}^{k} \frac{1}{k} \E_{(x, y) \sim {\mathcal{P}_{train_g}}}[\ell(\theta; (x, y))].
\end{align}

\paragraph{Domain Independent (DomainInd)} learning adjusts the model architecture by replacing the single classifier head with $k$ separate classifier heads, each corresponding to a subgroup, such that although each sample is passed through the same encoder, the decoder can be fine-tuned to each subgroup \cite{wangFairnessVisualRecognition2020a}.  At inference, the classifier head with the largest activation makes the final prediction. 

\paragraph{Conditional learning of fair representations (CFair)} aims to learn fair and robust representations of the target label independent of any subgroup information \cite{zhao2020conditionallearningfairrepresentations}. 
This is achieved by aligning conditional representations of samples from different subgroups. 

\subsubsection{Subgroup definition across these methods}

In this work, we distinguish between two categories of bias mitigation methods: reweighting (gDRO and resampling), and model-based methods (DomainInd and CFair). This is because of differences in how subgroups are generally defined. While in model-based methods subgroups are defined solely based on attribute(s) $A$, reweighting methods can additionally define subgroups over $\mathcal{A} \times \mathcal{Y}$. This is because the former methods perform best if each subgroup contains both positive and negative samples. We expand on this distinction and its implications in Appendix \ref{subsec: subgroup construction}.

\subsection{Problem setting}
\label{subsect: problem setting}

We frame the task according to the fairness paradigm described in \citet{jonesRethinkingFairRepresentation2024}, whereby the objective is to generalise from a biased training distribution to an unbiased testing distribution. Dataset bias can take many forms (as discussed in \citet{jonesCausalPerspectiveDataset2024a}), but here, we focus on bias arising from spurious correlations between certain attributes of the data and the class label $Y$ which disappear in the unbiased deployment setting. 
We define two binary attributes $A$ and $S$ whose information is encoded in $X$ in the form of latent features $X_A$ and $X_S$ respectively. Features relating to the true class $Y$ are represented as $X_Y$. 
In the unbiased test setting, attribute-related information is independent of $Y$, such that $P(Y) = P(Y\mid A) = P(Y\mid S)$. However, in $P_{train}$, mechanisms like data selection may lead to the violation of this independence. Causal graphs illustrating this scenario are shown in Figure \ref{fig:causal graphs and dist diagrams}. 

In particular, in this study, we consider an example of a spurious correlation between $A$ and $Y$. Furthermore, we consider that the spuriously correlated samples are unevenly distributed across the second attribute $S$. This is inspired from a real world medical imaging example from \citet{olesenSlicingBiasExplaining2024}. Specifically, they describe a chest X-ray diagnosis model which shows better performance in one sex subgroup (denoted here by $S$). Previous attempts to reduce this disparity, such as resampling of the data across $S$, and manipulation of sex-specific regions of the X-rays, were ineffective, leading to the hypothesis that sex-specific differences were not causing the disparity  \cite{wengAreSexbasedPhysiological2023}. Subsequent work revealed that men and women presented different proportions of chest drains and ECG wires, which the model used as spurious correlations to predict disease, and that, balancing the test data with respect to these artefacts (corresponding to $A$ in this work) resulted in equal performance across sexes. To mimic this example, we use a semi-synthetic setting where the subgroup $S=0$ contains 95\% spuriously correlated samples, while the subgroup $S=1$ contains a smaller, though still substantial, proportion of spuriously correlated samples (80\%). With this setup, there would be a substantial difference in a model's performance across subgroups if it only correctly classified samples satisfying this correlation. 

We represent each distribution as probability vectors in \( \mathbb{R}^8 \) with each element corresponding to the probability of sampling one $(Y,S,A)$ subgroup. The unbiased distribution is defined to be uniform, i.e., $\mathcal{P}_{\text{unbiased}} = \left[\frac{1}{8}, \frac{1}{8}, \ldots, \frac{1}{8} \right]$, while $\mathcal{P}_{\text{train}} = [
\frac{0.95}{4},\ 
\frac{0.05}{4},\ 
\frac{0.8}{4},\ 
\frac{0.2}{4},\ 
\frac{0.05}{4},\ 
\frac{0.95}{4},\ 
\frac{0.2}{4},\ 
\frac{0.8}{4}
]$. Further details are presented in Table \ref{tab: train test dist} and Figure \ref{fig:causal graphs and dist diagrams}.


\begin{table}[h]
\centering
\caption{Probability distributions across $Y$, $A$, and $S$ in the biased train and validation dataset and the unbiased test set.}
\label{tab: train test dist}
\begin{center}
\resizebox{\columnwidth}{!}{%
\begin{tabular}{lll}
\toprule
\textbf{Probability distributions}                                & $\mathbf{\mathcal{P}_{\text{train}}}$ & $\mathbf{\mathcal{P}_{\text{unbiased}}}$ \\ \midrule
\textbf{$P(Y=1)$}                          & 0.5                               & 0.5                        \\ 
\textbf{$P(A=1)$}                            & 0.5                               & 0.5                        \\ 
\textbf{$P(Y=0\mid S=0) = P(Y=0\mid S=1)$}         & 0.5                               & 0.5                        \\ 
\textbf{$P(Y=0\mid A=0) = P(Y=1\mid A=1)$}         & 0.875                              & 0.5                        \\ 
\textbf{$P(Y=0,A=0\mid S=0) = P(Y=1,A=1\mid S=0)$} & 0.95                              & 0.5                        \\ 
\textbf{$P(Y=0,A=0\mid S=1) = P(Y=1,A=1\mid S=1)$} & 0.8                               & 0.5                        \\ \bottomrule
\end{tabular}%
}
\end{center}
\end{table}

\section{Experiments}
\label{experiments}
\subsection{Subgroup generation}

To understand how different subgroup definitions affect the generalisation performance of bias mitigation methods, we construct multiple sets of subgroups based on $A$, $S$, and $Y$ and train each model with them. Our goal is to simulate realistic scenarios, for instance where one may only have access to certain variables or to noisy subgroup annotations, or when deciding between the use of coarse or fine-grained subgroups (e.g. a level of ethnicity categorisation, or discretising on a continuous variable like age). We denote subgroups constructed as the intersection of multiple variables $a \times b$ as $(a,b)$. 

For data-based methods, these include: 1) subgroups based on a single variable: $A$, $Y$, and $S$ 2) subgroups based on the intersection of two or three variables: $(A,Y)$, $(S,Y)$, and  $(Y,S,A)$ 3) SC/no-SC subgroups where we group the two bias-aligned $(A,Y)$ subgroups together and the two bias-conflicting $(A,Y)$ subgroups together 4) subgroups based on the random splitting of existing subgroups: $(A,Y)_8$, $(S,Y)_8$ which split each $(A,Y)$ and $(S,Y)$ subgroup in two such that there are 8 subgroups in total 5) 4 completely random subgroups. For model-based methods, we do not include subgroupings based on $Y$ and  also add $A_4$,  $S_4$ ($A$ and $S$ subgroups randomly split in two respectively), and $(A,S)$ subgroups. Finally,  we explore 6) the impact of noise in subgroup annotations by injecting noise (in the form of mislabelling) into 1 - 50\% of the $(A,Y)$ and $A$ subgroup annotations for data- and model-based methods respectively. This noise does not affect the class labels $Y$. For Civil\_comments, in addition to the synthetic granular subgroups, we directly explore the impact of granularity on real subgroups, as the dataset contains subgroup information of multiple hierarchy levels. 

In total, we consider 15 subgroup combinations for reweighting-based methods and 12 for model-based methods. We illustrate these subgroups in Figure A\ref{fig:subgroup construction diagram}. This comprehensive set of subgroup definitions allows us to systematically investigate the impact of subgroup choices on the effectiveness of bias mitigation methods.

\subsection{Datasets and tasks}
We evaluate performance in image classification tasks in four datasets which we construct to satisfy the distributions specified in Figure \ref{fig:causal graphs and dist diagrams}. We summarise all details in Table A\ref{tab: dataset details}.

We adapt the \textbf{MNIST} dataset \cite{lecunGradientbasedLearningApplied1998} by binarising the classification task into predicting whether a digit is even or odd ($Y$). We add additional attributes by modifying the image background colour ($A$) to black or white, and colouring the foreground ($S$) as red or green. This controlled setting allows for clear evaluation of subgroup influences in a simple task. 

To explore a more challenging and realistic setting, we repeat the experiments with chest X-ray images from the CheXpert dataset (\textbf{CXP}) \cite{irvinCheXpertLargeChest2019}. The task is classification of the presence of pleural effusion ($Y$). $S$ is the sex of the patient and $A$ the presence of a pacemaker, using annotations provided in \cite{AnthonyMahalnobis2023}. 

We explore another real vision dataset commonly used in fair ML research, \textbf{CelebA} \cite{liu2015faceattributes}. The task is binary classification of whether the individual has blonde hair ($Y$) with additional attributes $A$, perceived gender, and $S$, whether the individual is smiling. 

Finally, we explore whether our findings extend to the text modality through the use of the \textbf{Civil\_comments} dataset, also commonly used in fair ML \cite{civilcommentsdataset}. The target $Y$ is toxicity prediction, with $A$ being any mention of gender, and $S$ any mention of religion. Additionally, as this dataset contains multiple levels of subgroup annotations, we do a real experiment on the impact of subgroup granularity (instead of randomly splitting subgroups in two). We compare mitigation on the $A$ groups to mitigation on granular $A$ groups (e.g. any mention of males, any mention of another gender, and no mention of any gender), and likewise for $S$ (any mention of the Christian religion specifically).

\subsection{Implementation details}
We implement and train models with each of the gDRO, resampling, DomainInd, and CFair bias mitigation methods. We apply each method to each of our generated subgroups and average the results over three random seeds. We repeat this process for the four datasets, comparing performance of the bias mitigation methods with the baseline ERM method. 
In total, we train 306 models, with $\sim$40 NVIDIA A100 hours of compute. The training strategy, hyperparameters, architectures etc. are the same across all models, as detailed in Table A\ref{tab: implementation details}, except for  necessary adjustments to apply each bias mitigation method. The code is available \href{https://github.com/anissa218/subgroups_bias_mit}{here}.

We report the mean and standard deviation of the aggregate area under receiver operating characteristic curve (AUC) on the unbiased test set, alongside worst-group accuracy and accuracy gap across subgroups. We select these measures for their directness and simplicity compared to other fairness criteria. We do \textit{not} vary subgroup definition for evaluation and only evaluate accuracy with respect to $S$ and $A$ subgroups, as we are simply interested in how subgroup definition affects the mitigation process.

\section{Results and discussion}
\label{results_and_discussion}
\subsection{ERM performance drops on the unbiased test set}
\begin{table*}[ht]
\centering
\caption{The baseline model performance shows a sharp decrease on the unbiased test set for all datasets. Subgroup-wise accuracy also reveals large disparities with respect to $S$ in both datasets. Mean and standard deviation across three random seeds are shown.}
\resizebox{0.9\textwidth}{!}{%
\begin{tabular}{l|cc|cc|cc|cc}
\toprule
\multirow{2}{*}{\textbf{Accuracy}} & \multicolumn{2}{c|}{\textbf{MNIST}} & \multicolumn{2}{c|}{\textbf{CXP}} & \multicolumn{2}{c|}{\textbf{Civil\_comments}} & \multicolumn{2}{c}{\textbf{CelebA}} \\
                          & \textbf{Val}            & \textbf{Test}           & \textbf{Val}            & \textbf{Test}           & \textbf{Val}            & \textbf{Test}           & \textbf{Val}            & \textbf{Test}           \\ \midrule
\textbf{Overall}                   & 0.943 $\pm$ 0.012 & 0.698 $\pm$ 0.054 & 0.898 $\pm$ 0.007 & 0.659 $\pm$ 0.007 & 0.886 $\pm$ 0.003 & 0.726 $\pm$ 0.024 & 0.954 $\pm$ 0.003 & 0.865 $\pm$ 0.007 \\ \midrule
\textbf{Min ($A$)}                 & 0.936 $\pm$ 0.025 & 0.694 $\pm$ 0.003 & 0.869 $\pm$ 0.013 & 0.554 $\pm$ 0.020 & 0.878 $\pm$ 0.006 & 0.724 $\pm$ 0.037 & 0.952 $\pm$ 0.002 & 0.838 $\pm$ 0.019 \\
\textbf{Gap ($A$)  }               & 0.014 $\pm$ 0.026 & 0.008 $\pm$ 0.112 & 0.058 $\pm$ 0.014 & 0.220 $\pm$ 0.028 & 0.016 $\pm$ 0.013 & 0.004 $\pm$ 0.047 & 0.005 $\pm$ 0.007 & 0.055 $\pm$ 0.025 \\
\textbf{Min ($S$)}                 & 0.917 $\pm$ 0.024 & 0.599 $\pm$ 0.061 & 0.841 $\pm$ 0.010 & 0.633 $\pm$ 0.020 & 0.837 $\pm$ 0.002 & 0.726 $\pm$ 0.029 & 0.940 $\pm$ 0.004 & 0.861 $\pm$ 0.002 \\
\textbf{Gap ($S$)}                 & 0.052 $\pm$ 0.025 & 0.194 $\pm$ 0.080 & 0.116 $\pm$ 0.015 & 0.052 $\pm$ 0.026 & 0.095 $\pm$ 0.009 & 0.006 $\pm$ 0.047 & 0.029 $\pm$ 0.008 & 0.009 $\pm$ 0.014 \\
\bottomrule
\end{tabular}%
}
\label{tab:baseline_performance}
\end{table*}
The baseline model, trained with no bias mitigation, shows a sharp drop in performance when tested on the unbiased test set, with a decrease of 0.10 to 0.25 in AUC for all datasets (Table \ref{tab:baseline_performance}). This drop occurs because the model has learned to rely on the spurious attribute $A$ as a proxy for $Y$, and this correlation is absent in the unbiased test set. Moreover, the ERM model exhibits disparities in performance. This is particularly pronounced for the $S$ subgroups, both on the biased validation set and on the unbiased test set. A standard approach would therefore have been to apply mitigation to the $S$ or $(S,Y)$ subgroups. In the following sections, we explore whether various bias mitigation methods, used with different groupings, can improve test set performance and reduce these disparities.

\subsection{Groupings used for mitigation strongly impact bias mitigation performance}
\begin{figure*}[ht]
    \centering
{\includegraphics[height=7.8cm]{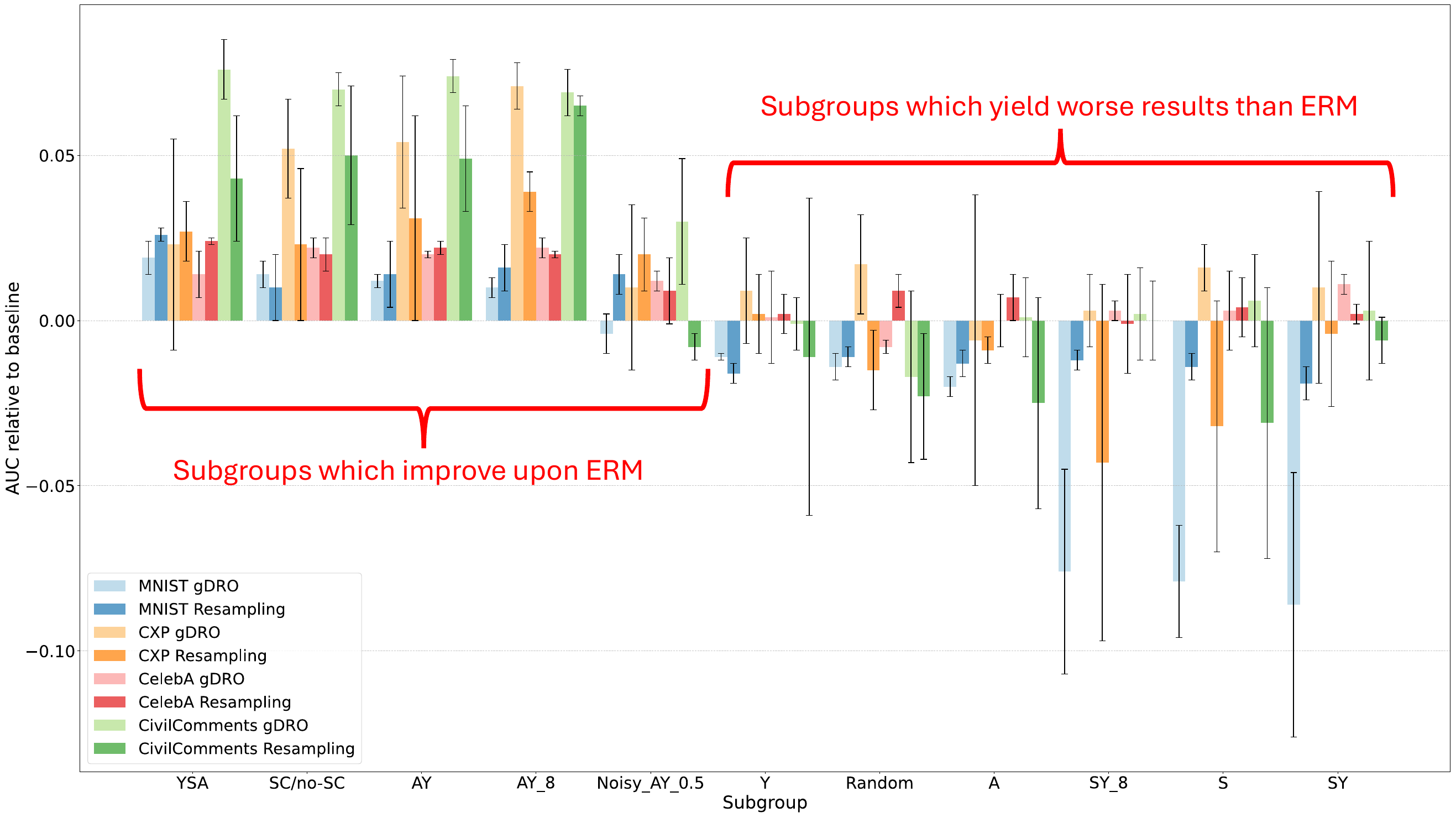}}
    \caption{Performance on the$\mathcal{P}_{unbiased}$ in gDRO and resampling is highly dependent on the subgroups used across all four datasets. Bars represent overall change in AUC relative to the ERM baseline, with error bars indicating the standard deviation across 3 random seeds.}
    \label{fig:mnist cxp gdro resampling auc}
\end{figure*}

We find that test set performance is highly dependent on the subgroups used for mitigation, with some subgroups boosting overall performance by more than 0.07, while others reduce it by up to 0.08 (Figures \ref{fig:mnist cxp gdro resampling auc} and A\ref{fig:mnist cxp domainind cfair auc}). 
We first note that the four datasets reveal similar patterns across groupings for reweighting-based bias mitigation methods and model-based methods, suggesting that there are universal trends relating to the bias setting which generalise outside of the specific dataset, task, and mitigation method. 
For reweighting-based methods, subgroups based on $(A,Y)$, i.e., $(A,Y)$, $(A,Y)_8$, $(Y,S,A)$, $(A,Y)$ with small levels of noise, and SC/no-SC subgroups, enable the model to focus on the $(A,Y)$ pairs without the spurious correlation during training. Therefore the model learns to predict $Y$ independently of $A$, leading to better generalisation performance. Conversely, subgroups which do not take $(A,Y)$ information into account tend to result in worse performance than the baseline model (for instance $S$,$(S,Y)$,$(S,Y)_8$,$Y$, $A$, and Random subgroups), as they fail to guide the model away from relying on the spurious attribute $A$. 
For model-based methods, a similar pattern is evident (Figure B\ref{fig:mnist cxp domainind cfair auc}); $A$ subgroups generally present the best performance (analogous to $(A,Y)$ in data-based methods). For instance, with $A$ subgroups in DomainInd, performance is increased by 0.07 in the CXP dataset. On the other hand, $S$ subgroups consistently harm performance, decreasing it by up to 0.14 for CFair in MNIST.


\paragraph{Applying bias mitigation with certain groupings can lead to worse outcomes than ERM.} For instance, $S$ subgroups are detrimental to generalisation performance in 10 out of 16 experiments, and have no effect in the remaining 4 (Figures \ref{fig:mnist cxp gdro resampling auc}, A\ref{fig:mnist cxp domainind cfair auc}). They are detrimental despite the fact that there is a substantial disparity in performance between them (Table \ref{tab:baseline_performance}). This observation suggests that in the absence of information about the underlying mechanism causing bias, it may be better to refrain from using mitigation methods altogether, even for simple approaches like data balancing. We believe that this finding may partly explain the failures of bias mitigation methods reported in recent studies \citep{zongMEDFAIRBENCHMARKINGFAIRNESS2023,zietlowLevelingComputerVision2022,chenComprehensiveEmpiricalStudy2023,shresthaInvestigationCriticalIssues2022a}, where inadequate subgroup definitions may have caused counter-productive bias mitigation.

\begin{figure}[h]
    \centering
{\includegraphics[height=4cm]{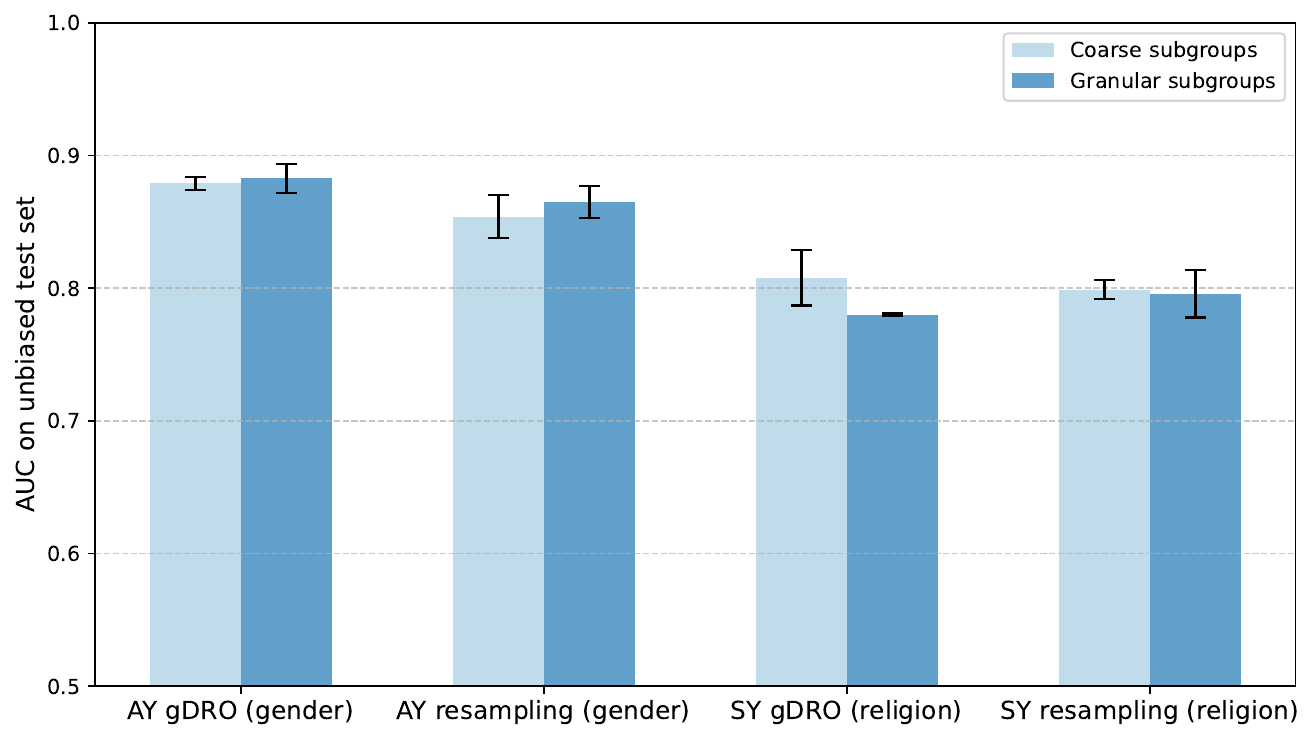}}
    \caption{Performance on $\mathcal{P}_{unbiased}$ is similar for the coarse and granular subgroups in Civil\_comments (mention of gender v.s. a specific male/female mention ($A$) and mention of any religion v.s. a specific mention of Christianity ($S$)). Error bars show standard deviation across 3 random seeds.}
    \label{fig:civil comments granularity plots}
\end{figure}

\paragraph{Increasing subgroup granularity has no significant effect on performance.} For example, there is little difference between results for $(A,Y)$ vs $(A,Y)_8$ and $(S,Y)$ vs $(S,Y)_8$ (Figure \ref{fig:mnist cxp gdro resampling auc}). This is most likely because the bias mitigation methods tested here are designed to maintain stability if there are many subgroups. In practice, this suggests that dividing subgroups into finer subgroups is unlikely to harm performance, and may even be beneficial if one is unsure which specific attributes may be responsible for underperformance. This insight could be relevant in settings where there are multiple levels of granularity to choose from (such as continuous data e.g., age, or multi-level categories e.g., ethnicity).
We validate this with the real coarse and granular subgroups in Civil\_comments, and indeed find no significant difference in unbiased generalisation for resampling and gDRO, as shown in Figure \ref{fig:civil comments granularity plots}.

\begin{figure}[htb]
    \centering
{\includegraphics[height=5cm]{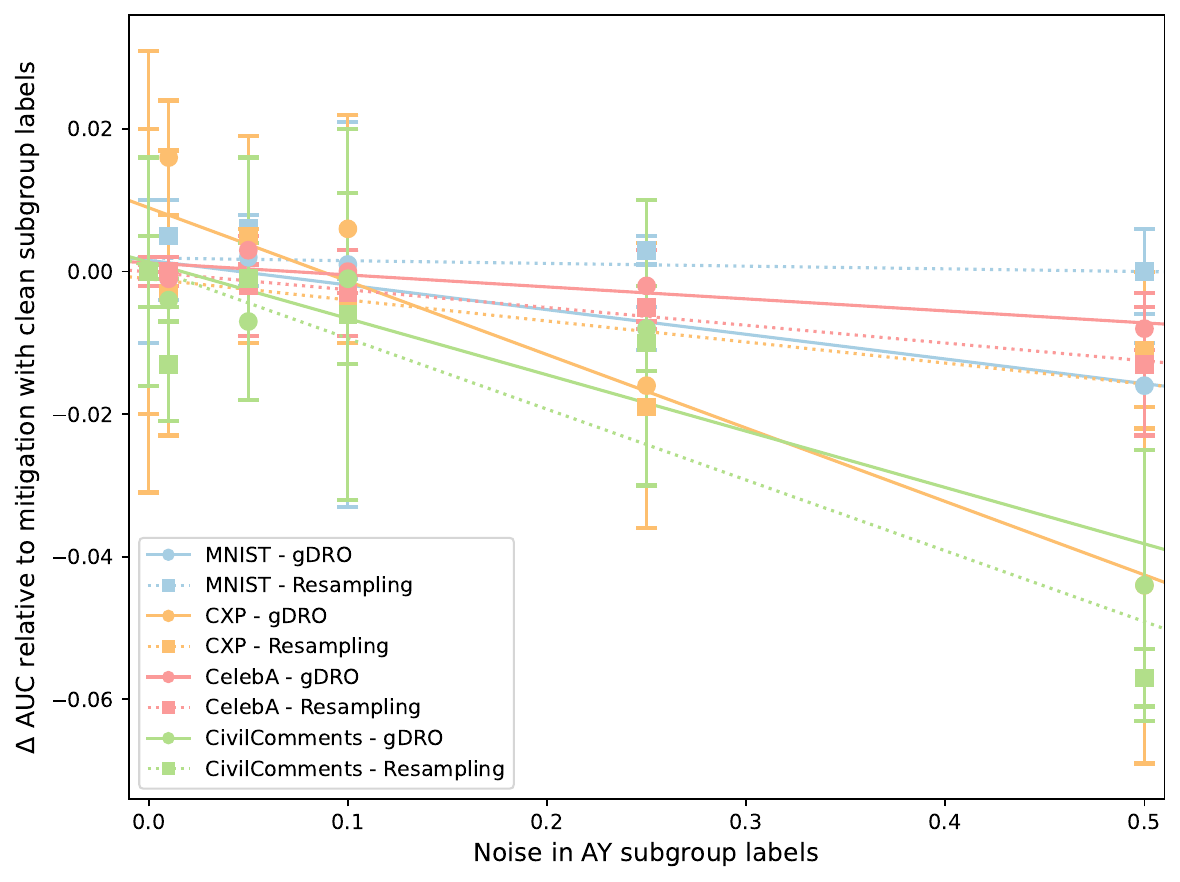}}
    \caption{Noise in $(A,Y)$ subgroup labels leads to a degradation in AUC for gDRO and resampling. Each dot represents mean performance on the unbiased test set for a specific grouping, with error bars indicating the standard deviation across 3 random seeds.}
    \label{fig:mnist cxp noise auc}
\end{figure}

\paragraph{Mitigation methods are relatively robust to noise.} As shown in Figure \ref{fig:mnist cxp noise auc}, performance of gDRO and resampling degrades when noise in the $(A,Y)$ annotations exceeds 10\%. However, even under these conditions, these methods still outperform the baseline model, indicating that they are relatively robust to annotation noise affecting a minority of subgroup annotations. This aligns with findings from \citet{awasthiEqualizedOddsPostprocessing2020} and \cite{stromberg2024for} who explore the impact of noise in post-processing and last-layer retraining respectively. A similar trend is observed for DomainInd in the CXP dataset, although the effect is less clear for other model-based experiments, where performance frequently falls below the baseline across all noise levels (Figure A\ref{fig:mnist cxp noise auc domainind resampling}). 

\paragraph{Random subgroups are generally detrimental.} Across the four datasets and mitigation methods we also find that using completely random subgroups generally worsens performance relative to ERM. This is most likely because the methods themselves are sub-optimal relative to ERM on the same distribution. This suggests that when one has no idea if the attributes for which they have annotations are related to possible biases, mitigation should not be conducted.

\subsection{The ability to recover the unbiased distribution is key to mitigation success}

We next aim to explain the observed variation in performance across different groupings. Inspired by \cite{zhouExaminingCombatingSpurious2021}, we explore whether it is possible to recover the unbiased distribution by weighting the chosen subgroups. Our hypothesis is that the closer the weighted $\mathcal{P}_{train}$, which we denote $\mathcal{P}_{\text{train}}^{w}$, is to the test distribution $\mathcal{P}_{unbiased}$, the better the model will perform on $\mathcal{P}_{unbiased}$. This reflects the broader consensus in the generalisation literature that aligning train and test distributions (e.g. through methods like data balancing) can reduce generalisation error in non-i.i.d. settings \cite{mansourdomainadaptation,dong2024doesdistributionmatchinghelp,wang2023causalbalancingdomaingeneralization}. For example, Ben-David et al. \cite{bendavidtheoryoflearning} show that for any hypothesis \( h \), assuming the same labelling function in both source and target domains, the test error is bounded as follows:
\[
\text{err}_{\text{test}}(h) \leq \text{err}_{\text{train}}(h) + D_{\text{TV}}(\mathcal{P}_{\text{train}}, \mathcal{P}_{\text{test}})
\]
where \( D_{\text{TV}}(\cdot, \cdot) \) denotes the total variation divergence.

We therefore choose to measure the divergence between $\mathcal{P}_{\text{train}}^{w}$ and  $\mathcal{P}_{unbiased}$ for each subgrouping to see whether this is a predictor of performance on the unbiased distribution. We model the divergence as a Kullback-Leibler divergence following prior work which give upper bounds \cite{aminian2024learningalgorithmgeneralizationerror,masihalearningunderdistributionmismatch,wu2024generalizationtransferlearninginformationtheoretic,nguyen2022klguideddomainadaptation} and lower bounds \cite{masihalearningunderdistributionmismatch} for expected generalisation error. In our case, Pinkser’s inequality implies that test error can be bounded as:
\[
\text{err}_{\text{unbiased}}(h) \leq \text{err}_{\text{train}}(h) + \sqrt{\frac{1}{2} \mathrm{KL}(\mathcal{P}_{\text{train}}^{w} \parallel \mathcal{P}_{\text{unbiased}})}.
\]
Since all our models reach a similarly low train error, we posit that the differences in upper bound are largely driven by the divergence between both distributions\footnote{We would ideally like to directly estimate generalisation under distribution shift (instead of just having an upper bound), but this would require very strong assumptions \cite{chuangestimatinggeneralization}.}. We explore whether the divergences achieved for each subgrouping correlate with generalisation error.

We assume that the difference between both distributions is attributable to differences in probabilities of sampling each $(Y,S,A)$ subgroup. Thus we represent each $\mathcal{P}_{\text{train}}^{w}$ as probability vectors in \( \mathbb{R}^8 \), as defined in Section \ref{subsect: problem setting}. This yields an initial KL divergence \( \mathrm{KL}(\mathcal{P}_{\text{train}} \parallel \mathcal{P}_{\text{unbiased}}) \approx 0.527 \). We next explore whether, for each possible subgrouping, by either resampling or gDRO, this divergence can be reduced. For resampling, each subgroup in the chosen subgrouping is resampled such that they are uniformly distributed. For instance, resampling across $(Y,S,A)$ would give $\mathcal{P}_{\text{train}}^{w} = \left[\frac{1}{8}, \frac{1}{8}, \ldots, \frac{1}{8} \right]$. For gDRO, the distribution of the subgroups is learned during training by determining what weights to apply to each subgroup. We determine the theoretically optimal weights $w$ by solving a convex optimisation problem:
\[
\min_{w \in \Delta^m} \ \mathrm{KL}(\mathcal{P}_{\text{train}}^{w} \parallel \mathcal{P}_{\text{unbiased}}),
\]
where \( \Delta^k \) denotes the probability simplex over the selected \( k \) subgroups.
We present the divergences obtained for all groupings in Table \ref{tab:p_test distance} with full explanations and calculations provided in the Appendix \ref{sec: kl div calculations}.

\begin{table}[htb]
\caption{Minimum $\mathrm{KL}(\mathcal{P}_{\text{train}}^{w} \parallel \mathcal{P}_{\text{unbiased}})$ achievable by reweighting subgroups with gDRO and resampling.}
\label{tab:p_test distance}
\renewcommand{\arraystretch}{0.5}
\begin{small}
\begin{center}
\begin{tabular}{@{}c|cc@{}}
\toprule
 & \multicolumn{2}{c}{\textbf{KL divergence to }$P_{unbiased}$} \\
 \textbf{Grouping}     & \textbf{gDRO} & \textbf{Resampling} \\ \midrule
A                      & 0.527                 & 0.527                      \\
Y                      & 0.527                 & 0.527                      \\
S                      & 0.527                 & 0.527                      \\
AY                     & 0.113                 & 0.113                      \\
SY                     & 0.527                 & 0.527                      \\
YSA                    & 0.000                 & 0.000                      \\
SC/no-SC               & 0.113                 & 0.113                      \\
AY$_8$                 & 0.113                 & 0.113                      \\
SY$_8$                 & 0.527                 & 0.527                      \\
Random                 & 0.527                 & 0.527                      \\
Noisy\_{${AY_{0.01}}$} & 0.113                 & 0.113                      \\
Noisy\_{${AY_{0.05}}$} & 0.113                 & 0.114                      \\
Noisy\_{${AY_{0.10}}$} & 0.113                 & 0.116                      \\
Noisy\_{${AY_{0.25}}$} & 0.114                 & 0.131                      \\
Noisy\_{${AY_{0.50}}$} & 0.118                 & 0.189                      \\
\bottomrule
\end{tabular}
\end{center}
\end{small}
\end{table}

\begin{figure*}[htb]
\centering
{\includegraphics[height=5cm]{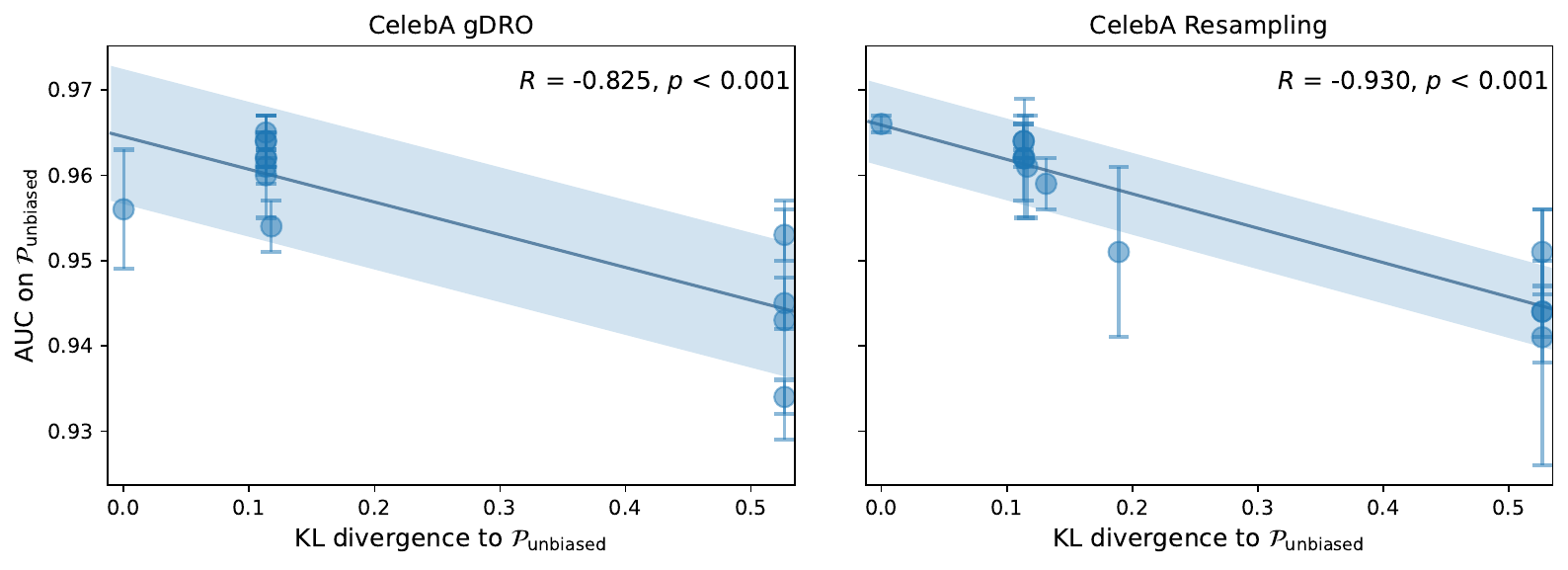}}
\caption{The ability to recover the unbiased test distribution ($\mathcal{P}_{unbiased}$) is a significant predictor of overall generalisation performance on $\mathcal{P}_{unbiased}$ for gDRO and resampling. Each dot represents mean AUC on the unbiased test set for a specific grouping, with error bars indicating the standard deviation across 3 random seeds. Pearson's correlation coefficients and associated p-values are also shown.}
\label{fig:mnist p_test distance auc}
\end{figure*}

We observe a high correlation (between 0.73 and 0.97) between the minimum achievable divergence and the performance of each model across the four datasets for resampling and gDRO respectively (Figures \ref{fig:mnist p_test distance auc}, F\ref{fig:kl div auc all datasets}). This aligns with our hypothesis that the extent to which the unbiased distribution can be restored during training significantly influences generalisation performance. Of course, other factors may still impact performance, such as inherent differences in task difficulty across subgroups \cite{PETERSEN2023100790}. We also note that we are only able to do this analysis by assuming that any divergence between distributions is fully attributable to differences in $P(Y,S,A)$ and because we have full knowledge of how these distributions change at test time. These assumptions would rarely be validated in practical settings. Despite this, our results suggest that assessing whether an unbiased distribution can be recovered provides a useful starting point for defining subgroups in bias mitigation. Thus, \textbf{defining subgroups based on the cause of generalisation error may be more effective than simply defining subgroups based on observed disparities}.

These divergences also explain previous observations, such as the similarity of results when subgroup granularity increases (KL divergence is also unchanged), robustness to noise (KL divergences show little change), and the similarity between results for gDRO and resampling (also similar divergences). Moreover, it is interesting to note that incorporating $S$ into the $(A,Y)$ groups (to get $(Y,S,A)$) is the most optimal grouping, as we observe empirically for MNIST (Figure \ref{fig:mnist cxp gdro resampling auc}). Although $S$ is not involved in the $(A,Y)$ spurious correlation and not a cause for poor generalisation performance, simply reweighting the four $(A,Y)$ groups induces an unwanted correlation between $S$ and $Y$, since the $(A,Y)$ groups are imbalanced with respect to $S$ (Table~\ref{tab: train test dist}).

\subsection{Subgroup choice also impacts disparities}
In addition to overall generalisation performance, we explore the impact of subgroup choice on disparities, specifically between $S$ subgroups (Table \ref{tab:gdro_resampling_s_acc} and \hyperref[tab:s disparities full table]{D9}). Fairness results largely align with overall generalisation performance. We make the surprising observation that the best results in terms of disparities for a given grouping are not necessarily achieved by using that grouping in the bias mitigation process. For example, $(A,Y)$ subgroups show far better worst-group-accuracy and smaller accuracy gaps across $S$ subgroups than when the $S$ attribute is used for grouping (e.g. $S$, $SY$, $SY_8$) for both datasets and methods. The minimum accuracy is up to 0.21 higher and an accuracy gap up to 10 times smaller. This is because the $(A,Y)$ spurious correlation is causing the disparity in performance across $S$, and it can only be unlearned with certain groupings. This finding contradicts the general assumption that in order to improve fairness with respect to a certain subgroup, this subgroup should be used in the bias mitigation process \cite{mehrabisurveyfairness2021}. To clarify, we do not advocate using different subgroups for evaluation, but rather that alternative subgroup definitions in \textit{mitigation} may better improve fairness.
\begin{table}[htb]
\begin{small}
\begin{center}
\caption{Worst group accuracy and accuracy gap across S groups, averaged for gDRO and resampling. While some groupings lead to improved fairness relative to the baseline (green), others are detrimental (red). Most groupings involving $S$ have the opposite of their intended effect and decrease fairness with respect to $S$. We report mean and standard deviation across three random seeds.}
\label{tab:gdro_resampling_s_acc}
\begin{tabular}{lll}
\toprule
\textbf{Grouping} &        \textbf{min Acc.} &        \textbf{Acc. gap} \\
\midrule
\textbf{baseline}     &  0.705 $\pm$ 0.025 &  0.064 $\pm$ 0.045 \\
\rowcolor{red!20}\textbf{A}            &  0.696 $\pm$ 0.019 &  0.074 $\pm$ 0.034 \\
\rowcolor{red!20}\textbf{Y}            &  0.696 $\pm$ 0.022 &  0.077 $\pm$ 0.037 \\
\rowcolor{red!20}\textbf{S}            &  0.694 $\pm$ 0.024 &  0.072 $\pm$ 0.039 \\
\rowcolor{green!20}\textbf{AY}           &   0.779 $\pm$ 0.02 &   0.032 $\pm$ 0.03 \\
\rowcolor{red!20}\textbf{SY}           &  0.699 $\pm$ 0.016 &  0.074 $\pm$ 0.026 \\
\rowcolor{green!20}\textbf{YSA}          &  0.792 $\pm$ 0.013 &   0.02 $\pm$ 0.022 \\
\rowcolor{green!20}\textbf{SC/no-SC}     &   0.772 $\pm$ 0.02 &    0.04 $\pm$ 0.03 \\
\rowcolor{green!20}$\mathbf{AY_8}$         &  0.778 $\pm$ 0.022 &  0.041 $\pm$ 0.032 \\
\rowcolor{red!20}$\mathbf{SY_8}$      &    0.7 $\pm$ 0.021 &  0.072 $\pm$ 0.032 \\
\rowcolor{red!20}\textbf{Random}       &   0.693 $\pm$ 0.02 &  0.075 $\pm$ 0.035 \\
\rowcolor{green!20}\textbf{Noisy\_{$\mathbf{AY_{0.50}}$}} &  0.741 $\pm$ 0.016 &  0.056 $\pm$ 0.029 \\
\bottomrule
\end{tabular}
\end{center}
\end{small}
\end{table}

\subsection{Subgroup choice is similarly impactful in other settings}

Finally, we conduct various additional experiments to verify that our results hold in less restrictive settings. We implement Just-train-twice (JTT), a method which does not require subgroup labels at training, but still requires some for model selection, and find that, again, its success is dependent on subgroup choice (Appendix \ref{subsec: jtt}: Table C\ref{tab: jtt} and Figure C\ref{fig: full graph with jtt}). We also repeat the MNIST experiments in a setting where the spurious correlation is weaker such that overall there are 77.5\% spuriously correlated samples (instead of 87.5\%). We find that while all results are higher overall, the same trends still appear, as shown in Figure G\ref{fig:mnist 85_70 p_test distance auc}. We also explore whether the relatively small size of the datasets we use (as we are constrained by the availability of each $(Y,S,A)$ combination) might impact our findings by comparing our results on the full 60K MNIST dataset to a downsampled 7K version. We find largely similar trends, with an example shown in Figure G\ref{fig:mnist 7k 60k kl div auc}.

\subsection{Limitations and future work}
This work is limited to a specific setting of bias i.e. spurious correlations, but many other types exist, including bias caused by differences in the manifestation of $Y$ across subgroups, differences in annotation of $Y$ across subgroups, and under-representation of certain subgroups. We hope that our research provides a solid foundation for further exploration into how subgroups should be defined for bias mitigation across a wide range of real bias settings. Another limitation is that, in practice, outside of (semi-)synthetic scenarios like ours, one often lacks extra annotations on other attributes, and is not perfectly aware of what might be causing underperformance. While we advocate for data collectors to gather as much metadata as possible to enable precise analyses, we recognise that this is not always feasible. Therefore, we encourage practitioners to carefully analyse their models' errors and thoughtfully investigate potential causes of underperformance before implementing bias mitigation strategies for specific subgroups.

\section{Conclusions}
\label{conclusions}
To our knowledge, this is the first work to specifically and comprehensively consider how subgroup definition can impact existing bias mitigation methods. We demonstrate the extent to which certain subgroup definitions can ``make or break'' bias mitigation methods, and provide an explanation as to why this occurs. We urge practitioners to carefully consider possible causes of bias rather than indiscriminately applying bias mitigation techniques to any underperforming group. Our work enables more consistent and effective bias mitigation in real-world applications.

\section*{Acknowledgements}

A.A. is supported by the EPSRC grant number EP/S024093/1, the Centre for Doctoral Training SABS: R3, University of Oxford, and by GE Healthcare. C.J. is supported by Microsoft Research, EPSRC, and The Alan Turing Institute through a Microsoft PhD scholarship and a Turing PhD enrichment award. B.G. received support from the Royal Academy of Engineering as part of his Kheiron/RAEng Research Chair.
The computational aspects of this research were supported by the Wellcome Trust Core Award Grant Number 203141/Z/16/Z and the NIHR Oxford BRC. The views expressed are those of the author(s) and not necessarily those of the NHS, the NIHR or the Department of Health.
The authors would also like to thank the reviewers of this paper whose comments contributed to substantially improving the work.

\section*{Impact Statement}

This paper presents work whose goal is to advance the field of Machine Learning. There are many potential societal consequences of our work, none which we feel must be specifically highlighted here.

\bibliography{MyLibrary}

\begin{thebibliography}{62}
\providecommand{\natexlab}[1]{#1}
\providecommand{\url}[1]{\texttt{#1}}
\expandafter\ifx\csname urlstyle\endcsname\relax
  \providecommand{\doi}[1]{doi: #1}\else
  \providecommand{\doi}{doi: \begingroup \urlstyle{rm}\Url}\fi

\bibitem[Ahn et~al.(2022)Ahn, Kim, and Yun]{ahnMitigatingDatasetBias2022}
Ahn, S., Kim, S., and Yun, S.-Y.
\newblock Mitigating dataset bias by using per-sample gradient.
\newblock In \emph{NeurIPS 2022 Workshop on Distribution Shifts: Connecting Methods and Applications}, 2022.
\newblock URL \url{https://openreview.net/forum?id=ihlU5X4SEE}.

\bibitem[Alloula et~al.(2024)Alloula, Mustafa, McGowan, and Papie{\.{z}}]{10.1007/978-3-031-72787-0_14}
Alloula, A., Mustafa, R., McGowan, D.~R., and Papie{\.{z}}, B.~W.
\newblock On biases in a uk biobank-based retinal image classification model.
\newblock In Puyol-Ant{\'o}n, E., Zamzmi, G., Feragen, A., King, A.~P., Cheplygina, V., Ganz-Benjaminsen, M., Ferrante, E., Glocker, B., Petersen, E., Baxter, J. S.~H., Rekik, I., and Eagleson, R. (eds.), \emph{Ethics and Fairness in Medical Imaging}, pp.\  140--150, Cham, 2024. Springer Nature Switzerland.
\newblock ISBN 978-3-031-72787-0.

\bibitem[Aminian et~al.(2024)Aminian, Masiha, Toni, and Rodrigues]{aminian2024learningalgorithmgeneralizationerror}
Aminian, G., Masiha, S., Toni, L., and Rodrigues, M. R.~D.
\newblock Learning algorithm generalization error bounds via auxiliary distributions, 2024.
\newblock URL \url{https://arxiv.org/abs/2210.00483}.

\bibitem[Anthony \& Kamnitsas(2023)Anthony and Kamnitsas]{AnthonyMahalnobis2023}
Anthony, H. and Kamnitsas, K.
\newblock On the use of mahalanobis distance for out-of-distribution detection with neural networks for medical imaging.
\newblock In \emph{Uncertainty for Safe Utilization of Machine Learning in Medical Imaging}, pp.\  136--146. Springer Nature Switzerland, 2023.
\newblock \doi{10.1007/978-3-031-44336-7_14}.
\newblock URL \url{https://doi.org/10.1007%2F978-3-031-44336-7_14}.

\bibitem[Awasthi et~al.(2020)Awasthi, Kleindessner, and Morgenstern]{awasthiEqualizedOddsPostprocessing2020}
Awasthi, P., Kleindessner, M., and Morgenstern, J.
\newblock Equalized odds postprocessing under imperfect group information.
\newblock In \emph{Proceedings of the {{Twenty Third International Conference}} on {{Artificial Intelligence}} and {{Statistics}}}, pp.\  1770--1780. PMLR, June 2020.

\bibitem[Bayasi et~al.(2024)Bayasi, Fayyad, Bissoto, Hamarneh, and Garbi]{bayasiBiasPrunerDebiasedContinual2024}
Bayasi, N., Fayyad, J., Bissoto, A., Hamarneh, G., and Garbi, R.
\newblock {{BiasPruner}}: {{Debiased Continual Learning}} for {{Medical Image Classification}}.
\newblock In \emph{MICCAI}, 2024.

\bibitem[Ben-David et~al.(2010)Ben-David, Blitzer, Crammer, Kulesza, Pereira, and Vaughan]{bendavidtheoryoflearning}
Ben-David, S., Blitzer, J., Crammer, K., Kulesza, A., Pereira, F., and Vaughan, J.
\newblock A theory of learning from different domains.
\newblock \emph{Machine Learning}, 79:\penalty0 151--175, 2010.
\newblock URL \url{http://www.springerlink.com/content/q6qk230685577n52/}.

\bibitem[Borkan et~al.(2019)Borkan, Dixon, Sorensen, Thain, and Vasserman]{civilcommentsdataset}
Borkan, D., Dixon, L., Sorensen, J., Thain, N., and Vasserman, L.
\newblock Nuanced metrics for measuring unintended bias with real data for text classification.
\newblock \emph{CoRR}, abs/1903.04561, 2019.
\newblock URL \url{http://arxiv.org/abs/1903.04561}.

\bibitem[Buolamwini \& Gebru(2018)Buolamwini and Gebru]{buolamwiniGenderShadesIntersectional2018}
Buolamwini, J. and Gebru, T.
\newblock Gender {{Shades}}: {{Intersectional Accuracy Disparities}} in {{Commercial Gender Classification}}.
\newblock In \emph{Proceedings of the 1st {{Conference}} on {{Fairness}}, {{Accountability}} and {{Transparency}}}, pp.\  77--91. PMLR, January 2018.

\bibitem[Chen et~al.(2023)Chen, Zhang, Sarro, and Harman]{chenComprehensiveEmpiricalStudy2023}
Chen, Z., Zhang, J.~M., Sarro, F., and Harman, M.
\newblock A comprehensive empirical study of bias mitigation methods for machine learning classifiers.
\newblock \emph{ACM Trans. Softw. Eng. Methodol.}, 32\penalty0 (4), May 2023.
\newblock ISSN 1049-331X.
\newblock \doi{10.1145/3583561}.
\newblock URL \url{https://doi.org/10.1145/3583561}.

\bibitem[Chuang et~al.(2020)Chuang, Torralba, and Jegelka]{chuangestimatinggeneralization}
Chuang, C.-Y., Torralba, A., and Jegelka, S.
\newblock Estimating generalization under distribution shifts via domain-invariant representations.
\newblock In \emph{Proceedings of the 37th International Conference on Machine Learning}, ICML'20. JMLR.org, 2020.

\bibitem[Deng et~al.(2009)Deng, Dong, Socher, Li, Li, and Fei-Fei]{dengimagenet2009}
Deng, J., Dong, W., Socher, R., Li, L.-J., Li, K., and Fei-Fei, L.
\newblock Imagenet: A large-scale hierarchical image database.
\newblock In \emph{2009 IEEE Conference on Computer Vision and Pattern Recognition}, pp.\  248--255, 2009.
\newblock \doi{10.1109/CVPR.2009.5206848}.

\bibitem[Devlin et~al.(2018)Devlin, Chang, Lee, and Toutanova]{bertclassifier}
Devlin, J., Chang, M., Lee, K., and Toutanova, K.
\newblock {BERT:} pre-training of deep bidirectional transformers for language understanding.
\newblock \emph{CoRR}, abs/1810.04805, 2018.
\newblock URL \url{http://arxiv.org/abs/1810.04805}.

\bibitem[Dong et~al.(2024)Dong, Gong, Chen, Song, Zhang, and Li]{dong2024doesdistributionmatchinghelp}
Dong, Y., Gong, T., Chen, H., Song, S., Zhang, W., and Li, C.
\newblock How does distribution matching help domain generalization: An information-theoretic analysis, 2024.
\newblock URL \url{https://arxiv.org/abs/2406.09745}.

\bibitem[Dwork et~al.(2012)Dwork, Hardt, Pitassi, Reingold, and Zemel]{dworkFairnessAwareness2012}
Dwork, C., Hardt, M., Pitassi, T., Reingold, O., and Zemel, R.
\newblock Fairness through awareness.
\newblock In \emph{Proceedings of the 3rd {{Innovations}} in {{Theoretical Computer Science Conference}}}, {{ITCS}} '12, pp.\  214--226, New York, NY, USA, January 2012. Association for Computing Machinery.
\newblock ISBN 978-1-4503-1115-1.
\newblock \doi{10.1145/2090236.2090255}.

\bibitem[Ganin \& Lempitsky(2015)Ganin and Lempitsky]{ganinunsuperviseddomainadapt2015}
Ganin, Y. and Lempitsky, V.
\newblock Unsupervised domain adaptation by backpropagation.
\newblock In Bach, F. and Blei, D. (eds.), \emph{Proceedings of the 32nd International Conference on Machine Learning}, volume~37 of \emph{Proceedings of Machine Learning Research}, pp.\  1180--1189, Lille, France, 07--09 Jul 2015. PMLR.
\newblock URL \url{https://proceedings.mlr.press/v37/ganin15.html}.

\bibitem[Ghosal \& Li(2023)Ghosal and Li]{ghosalDistributionallyRobustOptimization2023}
Ghosal, S.~S. and Li, Y.
\newblock Distributionally {{Robust Optimization}} with {{Probabilistic Group}}.
\newblock \emph{Proceedings of the AAAI Conference on Artificial Intelligence}, 37\penalty0 (10):\penalty0 11809--11817, June 2023.
\newblock ISSN 2374-3468.
\newblock \doi{10.1609/aaai.v37i10.26394}.

\bibitem[Han \& Zou(2024)Han and Zou]{hanImprovingGroupRobustness2024}
Han, Y. and Zou, D.
\newblock Improving group robustness on spurious correlation requires preciser group inference.
\newblock In \emph{Forty-first International Conference on Machine Learning}, 2024.
\newblock URL \url{https://openreview.net/forum?id=KycvgOCBBR}.

\bibitem[He et~al.(2016)He, Zhang, Ren, and Sun]{he2016deep}
He, K., Zhang, X., Ren, S., and Sun, J.
\newblock Deep residual learning for image recognition.
\newblock In \emph{Proceedings of the IEEE conference on computer vision and pattern recognition}, pp.\  770--778, 2016.

\bibitem[Hebert-Johnson et~al.(2018)Hebert-Johnson, Kim, Reingold, and Rothblum]{pmlr-v80-hebert-johnson18a}
Hebert-Johnson, U., Kim, M., Reingold, O., and Rothblum, G.
\newblock Multicalibration: Calibration for the ({C}omputationally-identifiable) masses.
\newblock In Dy, J. and Krause, A. (eds.), \emph{Proceedings of the 35th International Conference on Machine Learning}, volume~80 of \emph{Proceedings of Machine Learning Research}, pp.\  1939--1948. PMLR, 10--15 Jul 2018.
\newblock URL \url{https://proceedings.mlr.press/v80/hebert-johnson18a.html}.

\bibitem[Huang et~al.(2017)Huang, Liu, Van Der~Maaten, and Weinberger]{huang2018denselyconnectedconvolutionalnetworks}
Huang, G., Liu, Z., Van Der~Maaten, L., and Weinberger, K.~Q.
\newblock Densely connected convolutional networks.
\newblock In \emph{2017 IEEE Conference on Computer Vision and Pattern Recognition (CVPR)}, pp.\  2261--2269, 2017.
\newblock \doi{10.1109/CVPR.2017.243}.

\bibitem[Idrissi et~al.(2022)Idrissi, Arjovsky, Pezeshki, and {Lopez-Paz}]{idrissiSimpleDataBalancing2022}
Idrissi, B.~Y., Arjovsky, M., Pezeshki, M., and {Lopez-Paz}, D.
\newblock Simple data balancing achieves competitive worst-group-accuracy.
\newblock In \emph{Proceedings of the {{First Conference}} on {{Causal Learning}} and {{Reasoning}}}, pp.\  336--351. PMLR, June 2022.

\bibitem[Irvin et~al.(2019)Irvin, Rajpurkar, Ko, Yu, Ciurea-Ilcus, Chute, Marklund, Haghgoo, Ball, Shpanskaya, Seekins, Mong, Halabi, Sandberg, Jones, Larson, Langlotz, Patel, Lungren, and Ng]{irvinCheXpertLargeChest2019}
Irvin, J., Rajpurkar, P., Ko, M., Yu, Y., Ciurea-Ilcus, S., Chute, C., Marklund, H., Haghgoo, B., Ball, R., Shpanskaya, K., Seekins, J., Mong, D.~A., Halabi, S.~S., Sandberg, J.~K., Jones, R., Larson, D.~B., Langlotz, C.~P., Patel, B.~N., Lungren, M.~P., and Ng, A.~Y.
\newblock Chexpert: a large chest radiograph dataset with uncertainty labels and expert comparison.
\newblock In \emph{Proceedings of the Thirty-Third AAAI Conference on Artificial Intelligence and Thirty-First Innovative Applications of Artificial Intelligence Conference and Ninth AAAI Symposium on Educational Advances in Artificial Intelligence}, AAAI'19/IAAI'19/EAAI'19. AAAI Press, 2019.
\newblock ISBN 978-1-57735-809-1.
\newblock \doi{10.1609/aaai.v33i01.3301590}.
\newblock URL \url{https://doi.org/10.1609/aaai.v33i01.3301590}.

\bibitem[Izmailov et~al.(2022)Izmailov, Kirichenko, Gruver, and Wilson]{izmailovfeaturelearning}
Izmailov, P., Kirichenko, P., Gruver, N., and Wilson, A.~G.
\newblock On feature learning in the presence of spurious correlations.
\newblock In \emph{Proceedings of the 36th International Conference on Neural Information Processing Systems}, NIPS '22, Red Hook, NY, USA, 2022. Curran Associates Inc.
\newblock ISBN 9781713871088.

\bibitem[Jain et~al.(2024)Jain, Hamidieh, Georgiev, Ilyas, Ghassemi, and Madry]{NEURIPS2024_abbbb25c}
Jain, S., Hamidieh, K., Georgiev, K., Ilyas, A., Ghassemi, M., and Madry, A.
\newblock Improving subgroup robustness via data selection.
\newblock In Globerson, A., Mackey, L., Belgrave, D., Fan, A., Paquet, U., Tomczak, J., and Zhang, C. (eds.), \emph{Advances in Neural Information Processing Systems}, volume~37, pp.\  94490--94511. Curran Associates, Inc., 2024.

\bibitem[Jones et~al.(2024)Jones, Castro, De~Sousa~Ribeiro, Oktay, McCradden, and Glocker]{jonesCausalPerspectiveDataset2024a}
Jones, C., Castro, D.~C., De~Sousa~Ribeiro, F., Oktay, O., McCradden, M., and Glocker, B.
\newblock A causal perspective on dataset bias in machine learning for medical imaging.
\newblock \emph{Nature Machine Intelligence}, 6\penalty0 (2):\penalty0 138--146, February 2024.
\newblock ISSN 2522-5839.
\newblock \doi{10.1038/s42256-024-00797-8}.

\bibitem[Jones et~al.(2025)Jones, Ribeiro, Roschewitz, Castro, and Glocker]{jonesRethinkingFairRepresentation2024}
Jones, C., Ribeiro, F. d.~S., Roschewitz, M., Castro, D.~C., and Glocker, B.
\newblock Rethinking fair representation learning for performance-sensitive tasks.
\newblock In \emph{The Thirteenth International Conference on Learning Representations}, 2025.
\newblock URL \url{https://openreview.net/forum?id=pBZntPrdrI}.

\bibitem[Kearns et~al.(2018)Kearns, Neel, Roth, and Wu]{kearnsPreventingFairnessGerrymandering2018}
Kearns, M., Neel, S., Roth, A., and Wu, Z.~S.
\newblock Preventing {{Fairness Gerrymandering}}: {{Auditing}} and {{Learning}} for {{Subgroup Fairness}}.
\newblock In \emph{Proceedings of the 35th {{International Conference}} on {{Machine Learning}}}, pp.\  2564--2572. PMLR, July 2018.

\bibitem[Kim et~al.(2024)Kim, Kang, Ahn, Ok, and Kwak]{kimImprovingRobustnessMultiple2024}
Kim, N., Kang, J., Ahn, S., Ok, J., and Kwak, S.
\newblock Improving {{Robustness}} to {{Multiple Spurious Correlations}} by {{Multi-Objective Optimization}}.
\newblock In \emph{Proceedings of the 41st {{International Conference}} on {{Machine Learning}}}, pp.\  24040--24058. PMLR, July 2024.

\bibitem[Kirichenko et~al.(2023)Kirichenko, Izmailov, and Wilson]{kirichenko2023layerretrainingsufficientrobustness}
Kirichenko, P., Izmailov, P., and Wilson, A.~G.
\newblock Last layer re-training is sufficient for robustness to spurious correlations.
\newblock In \emph{ICLR}, 2023.
\newblock URL \url{https://arxiv.org/abs/2204.02937}.

\bibitem[Krishnakumar et~al.(2021)Krishnakumar, Prabhu, Sudhakar, and Hoffman]{krishnakumarUDISUnsupervisedDiscovery}
Krishnakumar, A., Prabhu, V., Sudhakar, S., and Hoffman, J.
\newblock Udis: Unsupervised discovery of bias in deep visual recognition models.
\newblock In \emph{BMVC}, pp.\  143, 2021.
\newblock URL \url{https://www.bmvc2021-virtualconference.com/assets/papers/0362.pdf}.

\bibitem[Lecun et~al.(1998)Lecun, Bottou, Bengio, and Haffner]{lecunGradientbasedLearningApplied1998}
Lecun, Y., Bottou, L., Bengio, Y., and Haffner, P.
\newblock Gradient-based learning applied to document recognition.
\newblock \emph{Proceedings of the IEEE}, 86\penalty0 (11):\penalty0 2278--2324, November 1998.
\newblock ISSN 1558-2256.
\newblock \doi{10.1109/5.726791}.

\bibitem[Li et~al.(2023)Li, Evtimov, Gordo, Hazirbas, Hassner, Ferrer, Xu, and Ibrahim]{liWhacAMoleDilemmaShortcuts2023}
Li, Z., Evtimov, I., Gordo, A., Hazirbas, C., Hassner, T., Ferrer, C.~C., Xu, C., and Ibrahim, M.
\newblock A {{Whac-A-Mole Dilemma}}: {{Shortcuts Come}} in {{Multiples Where Mitigating One Amplifies Others}}.
\newblock \emph{2023 IEEE/CVF Conference on Computer Vision and Pattern Recognition (CVPR)}, pp.\  20071--20082, June 2023.
\newblock \doi{10.1109/CVPR52729.2023.01922}.

\bibitem[Liu et~al.(2021)Liu, Haghgoo, Chen, Raghunathan, Koh, Sagawa, Liang, and Finn]{liuJustTrainTwice2021}
Liu, E.~Z., Haghgoo, B., Chen, A.~S., Raghunathan, A., Koh, P.~W., Sagawa, S., Liang, P., and Finn, C.
\newblock Just train twice: Improving group robustness without training group information.
\newblock In Meila, M. and Zhang, T. (eds.), \emph{Proceedings of the 38th International Conference on Machine Learning}, volume 139 of \emph{Proceedings of Machine Learning Research}, pp.\  6781--6792. PMLR, 18--24 Jul 2021.
\newblock URL \url{https://proceedings.mlr.press/v139/liu21f.html}.

\bibitem[Liu et~al.(2015{\natexlab{a}})Liu, Luo, Wang, and Tang]{liu2015faceattributes}
Liu, Z., Luo, P., Wang, X., and Tang, X.
\newblock Deep learning face attributes in the wild.
\newblock In \emph{Proceedings of International Conference on Computer Vision (ICCV)}, December 2015{\natexlab{a}}.

\bibitem[Liu et~al.(2015{\natexlab{b}})Liu, Luo, Wang, and Tang]{liuDeepLearningFace}
Liu, Z., Luo, P., Wang, X., and Tang, X.
\newblock Deep learning face attributes in the wild.
\newblock In \emph{Proceedings of the 2015 IEEE International Conference on Computer Vision (ICCV)}, ICCV '15, pp.\  3730–3738, USA, 2015{\natexlab{b}}. IEEE Computer Society.
\newblock ISBN 9781467383912.
\newblock \doi{10.1109/ICCV.2015.425}.
\newblock URL \url{https://doi.org/10.1109/ICCV.2015.425}.

\bibitem[Mansour et~al.(2009)Mansour, Mohri, and Rostamizadeh]{mansourdomainadaptation}
Mansour, Y., Mohri, M., and Rostamizadeh, A.
\newblock Domain adaptation: Learning bounds and algorithms.
\newblock \emph{CoRR}, abs/0902.3430, 2009.
\newblock URL \url{http://arxiv.org/abs/0902.3430}.

\bibitem[Marani et~al.(2024)Marani, Hanini, Malayarukil, Christodoulidis, Vakalopoulou, and Ferrante]{maraniViGBiasVisuallyGrounded2024}
Marani, B.-E., Hanini, M., Malayarukil, N., Christodoulidis, S., Vakalopoulou, M., and Ferrante, E.
\newblock \emph{{{ViG-Bias}}: {{Visually Grounded Bias Discovery}} and {{Mitigation}}}, pp.\  414–429.
\newblock Springer Nature Switzerland, November 2024.
\newblock ISBN 9783031732027.
\newblock \doi{10.1007/978-3-031-73202-7_24}.
\newblock URL \url{http://dx.doi.org/10.1007/978-3-031-73202-7_24}.

\bibitem[Masiha et~al.(2021)Masiha, Gohari, Yassaee, and Aref]{masihalearningunderdistributionmismatch}
Masiha, M.~S., Gohari, A., Yassaee, M.~H., and Aref, M.~R.
\newblock Learning under distribution mismatch and model misspecification.
\newblock \emph{CoRR}, abs/2102.05695, 2021.
\newblock URL \url{https://arxiv.org/abs/2102.05695}.

\bibitem[Mehrabi et~al.(2021)Mehrabi, Morstatter, Saxena, Lerman, and Galstyan]{mehrabisurveyfairness2021}
Mehrabi, N., Morstatter, F., Saxena, N., Lerman, K., and Galstyan, A.
\newblock A survey on bias and fairness in machine learning.
\newblock \emph{ACM Comput. Surv.}, 54\penalty0 (6), July 2021.
\newblock ISSN 0360-0300.
\newblock \doi{10.1145/3457607}.
\newblock URL \url{https://doi.org/10.1145/3457607}.

\bibitem[Movva et~al.(2023)Movva, Shanmugam, Hou, Pathak, Guttag, Garg, and Pierson]{movvaCoarseRaceData2023}
Movva, R., Shanmugam, D., Hou, K., Pathak, P., Guttag, J., Garg, N., and Pierson, E.
\newblock Coarse race data conceals disparities in clinical risk score performance.
\newblock In \emph{Machine Learning for Healthcare Conference}, pp.\  443--472. PMLR, 2023.

\bibitem[Nguyen et~al.(2022)Nguyen, Tran, Gal, Torr, and Baydin]{nguyen2022klguideddomainadaptation}
Nguyen, A.~T., Tran, T., Gal, Y., Torr, P. H.~S., and Baydin, A.~G.
\newblock Kl guided domain adaptation, 2022.
\newblock URL \url{https://arxiv.org/abs/2106.07780}.

\bibitem[Olesen et~al.(2024)Olesen, Weng, Feragen, and Petersen]{olesenSlicingBiasExplaining2024}
Olesen, V., Weng, N., Feragen, A., and Petersen, E.
\newblock \emph{Slicing {{Through Bias}}: {{Explaining Performance Gaps}} in {{Medical Image Analysis}} Using {{Slice Discovery Methods}}}, pp.\  3–13.
\newblock Springer Nature Switzerland, October 2024.
\newblock ISBN 9783031727870.
\newblock \doi{10.1007/978-3-031-72787-0_1}.
\newblock URL \url{http://dx.doi.org/10.1007/978-3-031-72787-0_1}.

\bibitem[Park et~al.(2024)Park, Jung, Lee, Ye, and Lee]{parkSelfsupervisedDebiasingUsing2022}
Park, G.~Y., Jung, C., Lee, S., Ye, J.~C., and Lee, S.~W.
\newblock Self-supervised debiasing using low rank regularization.
\newblock In \emph{2024 {{IEEE}}/{{CVF Conference}} on {{Computer Vision}} and {{Pattern Recognition}} ({{CVPR}})}, 2024.
\newblock \doi{10.48550/ARXIV.2210.05248}.

\bibitem[Petersen et~al.(2023)Petersen, aPETERSEN2023100790nd Melanie~Ganz, and Feragen]{PETERSEN2023100790}
Petersen, E., aPETERSEN2023100790nd Melanie~Ganz, S.~H., and Feragen, A.
\newblock The path toward equal performance in medical machine learning.
\newblock \emph{Patterns}, 4\penalty0 (7):\penalty0 100790, 2023.
\newblock ISSN 2666-3899.
\newblock \doi{https://doi.org/10.1016/j.patter.2023.100790}.
\newblock URL \url{https://www.sciencedirect.com/science/article/pii/S2666389923001459}.

\bibitem[Pezeshki et~al.(2021)Pezeshki, Kaba, Bengio, Courville, Precup, and Lajoie]{pezeshkiGradientStarvationLearning2021}
Pezeshki, M., Kaba, S.-O., Bengio, Y., Courville, A., Precup, D., and Lajoie, G.
\newblock Gradient {{Starvation}}: {{A Learning Proclivity}} in {{Neural Networks}}, November 2021.

\bibitem[Pezeshki et~al.(2024)Pezeshki, Bouchacourt, Ibrahim, Ballas, Vincent, and Lopez-Paz]{pezeshki2024discovering}
Pezeshki, M., Bouchacourt, D., Ibrahim, M., Ballas, N., Vincent, P., and Lopez-Paz, D.
\newblock Discovering environments with {XRM}.
\newblock In \emph{ICML}, 2024.
\newblock URL \url{https://openreview.net/forum?id=IhWtRwIbos}.

\bibitem[Ricci~Lara et~al.(2022)Ricci~Lara, Echeveste, and Ferrante]{riccilaraAddressingFairnessArtificial2022}
Ricci~Lara, M.~A., Echeveste, R., and Ferrante, E.
\newblock Addressing fairness in artificial intelligence for medical imaging.
\newblock \emph{Nature Communications}, 13\penalty0 (1):\penalty0 4581, August 2022.
\newblock ISSN 2041-1723.
\newblock \doi{10.1038/s41467-022-32186-3}.

\bibitem[Sagawa* et~al.(2020)Sagawa*, Koh*, Hashimoto, and Liang]{sagawaDistributionallyRobustNeural2020}
Sagawa*, S., Koh*, P.~W., Hashimoto, T.~B., and Liang, P.
\newblock Distributionally robust neural networks.
\newblock In \emph{International Conference on Learning Representations}, 2020.
\newblock URL \url{https://openreview.net/forum?id=ryxGuJrFvS}.

\bibitem[Schrouff et~al.(2024)Schrouff, Bellot, Rannen-Triki, Malek, Albuquerque, Gretton, D'Amour, and Chiappa]{schrouffMindGraphWhen2024}
Schrouff, J., Bellot, A., Rannen-Triki, A., Malek, A., Albuquerque, I., Gretton, A., D'Amour, A.~N., and Chiappa, S.
\newblock Mind the graph when balancing data for fairness or robustness.
\newblock In \emph{The Thirty-eighth Annual Conference on Neural Information Processing Systems}, 2024.
\newblock URL \url{https://openreview.net/forum?id=LQR22jM5l3}.

\bibitem[Shrestha et~al.(2022)Shrestha, Kafle, and Kanan]{shresthaInvestigationCriticalIssues2022a}
Shrestha, R., Kafle, K., and Kanan, C.
\newblock An {{Investigation}} of {{Critical Issues}} in {{Bias Mitigation Techniques}}.
\newblock In \emph{2022 {{IEEE}}/{{CVF Winter Conference}} on {{Applications}} of {{Computer Vision}} ({{WACV}})}, pp.\  2512--2523, Waikoloa, HI, USA, January 2022. IEEE.
\newblock ISBN 978-1-66540-915-5.
\newblock \doi{10.1109/WACV51458.2022.00257}.

\bibitem[Stromberg et~al.(2024)Stromberg, Ayyagari, Welfert, Koyejo, Nock, and Sankar]{stromberg2024for}
Stromberg, N., Ayyagari, R., Welfert, M., Koyejo, S., Nock, R., and Sankar, L.
\newblock For robust worst-group accuracy, ignore group annotations.
\newblock \emph{Transactions on Machine Learning Research}, 2024.
\newblock ISSN 2835-8856.
\newblock URL \url{https://openreview.net/forum?id=l8E68fD6yp}.

\bibitem[Wang et~al.(2020{\natexlab{a}})Wang, Guo, Narasimhan, Cotter, Gupta, and Jordan]{wangRobustOptimizationFairness2020a}
Wang, S., Guo, W., Narasimhan, H., Cotter, A., Gupta, M., and Jordan, M.
\newblock Robust {{Optimization}} for {{Fairness}} with {{Noisy Protected Groups}}.
\newblock In \emph{Advances in {{Neural Information Processing Systems}}}, volume~33, pp.\  5190--5203. Curran Associates, Inc., 2020{\natexlab{a}}.

\bibitem[Wang et~al.(2023)Wang, Saxon, Li, Zhang, Zhang, and Wang]{wang2023causalbalancingdomaingeneralization}
Wang, X., Saxon, M., Li, J., Zhang, H., Zhang, K., and Wang, W.~Y.
\newblock Causal balancing for domain generalization.
\newblock In \emph{ICLR}, 2023.
\newblock URL \url{https://arxiv.org/abs/2206.05263}.

\bibitem[Wang et~al.(2020{\natexlab{b}})Wang, Qinami, Karakozis, Genova, Nair, Hata, and Russakovsky]{wangFairnessVisualRecognition2020a}
Wang, Z., Qinami, K., Karakozis, I.~C., Genova, K., Nair, P., Hata, K., and Russakovsky, O.
\newblock Towards {{Fairness}} in {{Visual Recognition}}: {{Effective Strategies}} for {{Bias Mitigation}}.
\newblock In \emph{2020 {{IEEE}}/{{CVF Conference}} on {{Computer Vision}} and {{Pattern Recognition}} ({{CVPR}})}, pp.\  8916--8925, Seattle, WA, USA, June 2020{\natexlab{b}}. IEEE.
\newblock ISBN 978-1-72817-168-5.
\newblock \doi{10.1109/CVPR42600.2020.00894}.

\bibitem[Weng et~al.(2023)Weng, Bigdeli, Petersen, and Feragen]{wengAreSexbasedPhysiological2023}
Weng, N., Bigdeli, S., Petersen, E., and Feragen, A.
\newblock Are sex-based physiological differences the cause of gender bias for chest x-ray diagnosis?
\newblock In \emph{Clinical Image-Based Procedures, Fairness of AI in Medical Imaging, and Ethical and Philosophical Issues in Medical Imaging: 12th International Workshop, CLIP 2023 1st International Workshop, FAIMI 2023 and 2nd International Workshop, EPIMI 2023 Vancouver, BC, Canada, October 8 and October 12, 2023 Proceedings}, pp.\  142–152, Berlin, Heidelberg, 2023. Springer-Verlag.
\newblock ISBN 978-3-031-45248-2.
\newblock \doi{10.1007/978-3-031-45249-9_14}.
\newblock URL \url{https://doi.org/10.1007/978-3-031-45249-9_14}.

\bibitem[Wu et~al.(2024)Wu, Manton, Aickelin, and Zhu]{wu2024generalizationtransferlearninginformationtheoretic}
Wu, X., Manton, J.~H., Aickelin, U., and Zhu, J.
\newblock On the generalization for transfer learning: An information-theoretic analysis, 2024.
\newblock URL \url{https://arxiv.org/abs/2207.05377}.

\bibitem[Xu et~al.(2024)Xu, Chen, Ling, Wang, and Shui]{xuIntersectionalUnfairnessDiscovery2024}
Xu, G., Chen, Q., Ling, C., Wang, B., and Shui, C.
\newblock Intersectional {{Unfairness Discovery}}.
\newblock In \emph{Proceedings of the 41st {{International Conference}} on {{Machine Learning}}}, pp.\  54888--54917. PMLR, July 2024.

\bibitem[Zhao et~al.(2020)Zhao, Coston, Adel, and Gordon]{zhao2020conditionallearningfairrepresentations}
Zhao, H., Coston, A., Adel, T., and Gordon, G.~J.
\newblock Conditional learning of fair representations.
\newblock In \emph{International Conference on Learning Representations}, 2020.
\newblock URL \url{https://openreview.net/forum?id=Hkekl0NFPr}.

\bibitem[Zhou et~al.(2021)Zhou, Ma, Michel, and Neubig]{zhouExaminingCombatingSpurious2021}
Zhou, C., Ma, X., Michel, P., and Neubig, G.
\newblock Examining and {{Combating Spurious Features}} under {{Distribution Shift}}.
\newblock In \emph{Proceedings of the 38th {{International Conference}} on {{Machine Learning}}}, pp.\  12857--12867. PMLR, July 2021.

\bibitem[Zietlow et~al.(2022)Zietlow, Lohaus, Balakrishnan, Kleindessner, Locatello, Scholkopf, and Russell]{zietlowLevelingComputerVision2022}
Zietlow, D., Lohaus, M., Balakrishnan, G., Kleindessner, M., Locatello, F., Scholkopf, B., and Russell, C.
\newblock Leveling down in computer vision: Pareto inefficiencies in fair deep classifiers.
\newblock \emph{2022 IEEE/CVF Conference on Computer Vision and Pattern Recognition (CVPR)}, pp.\  10400--10411, 2022.
\newblock URL \url{https://api.semanticscholar.org/CorpusID:247319023}.

\bibitem[Zong et~al.(2023)Zong, Yang, and Hospedales]{zongMEDFAIRBENCHMARKINGFAIRNESS2023}
Zong, Y., Yang, Y., and Hospedales, T.
\newblock Medfair: Benchmarking fairness for medical imaging.
\newblock In \emph{International Conference on Learning Representations (ICLR)}, 2023.

\end{thebibliography}
\bibliographystyle{icml2025}

\newpage
\appendix
\onecolumn
This appendix provides additional details and experiments that support the main text. It is structured as follows:

\begin{itemize}
    \item \ref{subsec: supplementary experimental details} Supplementary experimental details on the datasets used, subgroups constructed, and model implementations. 
    \item \ref{subsec: results more mitigation} Mitigation results for more bias mitigation methods: DomainInd and CFair.
    \item \ref{subsec: jtt} Mitigation without subgroup labels (Just Train Twice).
    \item \ref{subsec: supplementary results resampling and gdro} Supplementary results for gDRO and resampling.
    \item \ref{sec: kl div calculations} Explaining results through the divergence between $\mathcal{P}_{\text{train}}^{w}$ and $\mathcal{P}_{unbiased}$.
    \item \ref{subsec: correlation kl div generalisation} Correlation between KL divergence to the unbiased distribution and unbiased generalisation across all four datasets.
    \item \ref{subsec: ablations} Ablations on strength of SC and size of dataset.

\end{itemize}

\section{Supplementary experimental details}
\label{subsec: supplementary experimental details}

\subsection{Dataset details}

\begin{table}[h]
\caption{Details on the datasets used for mitigation experiments.}
\label{tab: dataset details}
\begin{small}
\begin{center}
\begin{tabular}{lllll}
\toprule
\textbf{Dataset}      & \textbf{MNIST}    & \textbf{CheXPert}       & \textbf{CelebA}  & \textbf{Civil\_comments} \\ \midrule
\textbf{Y}            & Even/odd digit    & Pleural effusion        & Blonde hair      & Toxicity                 \\ 
\textbf{A}            & Background colour & Presence of a pacemaker & Perceived gender & Gender                   \\ 
\textbf{S}            & Foreground colour & Sex                     & Smiling          & Religion                 \\ 
\textbf{Dataset size} & 60000              & 3225                    & 12500            & 8900                     \\ \bottomrule
\end{tabular}
\end{center}
\end{small}
\end{table}

We downsample some of the datasets from their original size because we are constrained by the availability of each $(Y,S,A)$ combination. For example, for CheXPert, pacemaker annotations are only available for 4862 images, and we have to further downsample the dataset to make it balanced with respect to disease ($Y$) and sex ($S$).

\subsection{Subgroup construction}
\label{subsec: subgroup construction}

\begin{figure*}[htb]
    \centering
{\includegraphics[width=\textwidth]{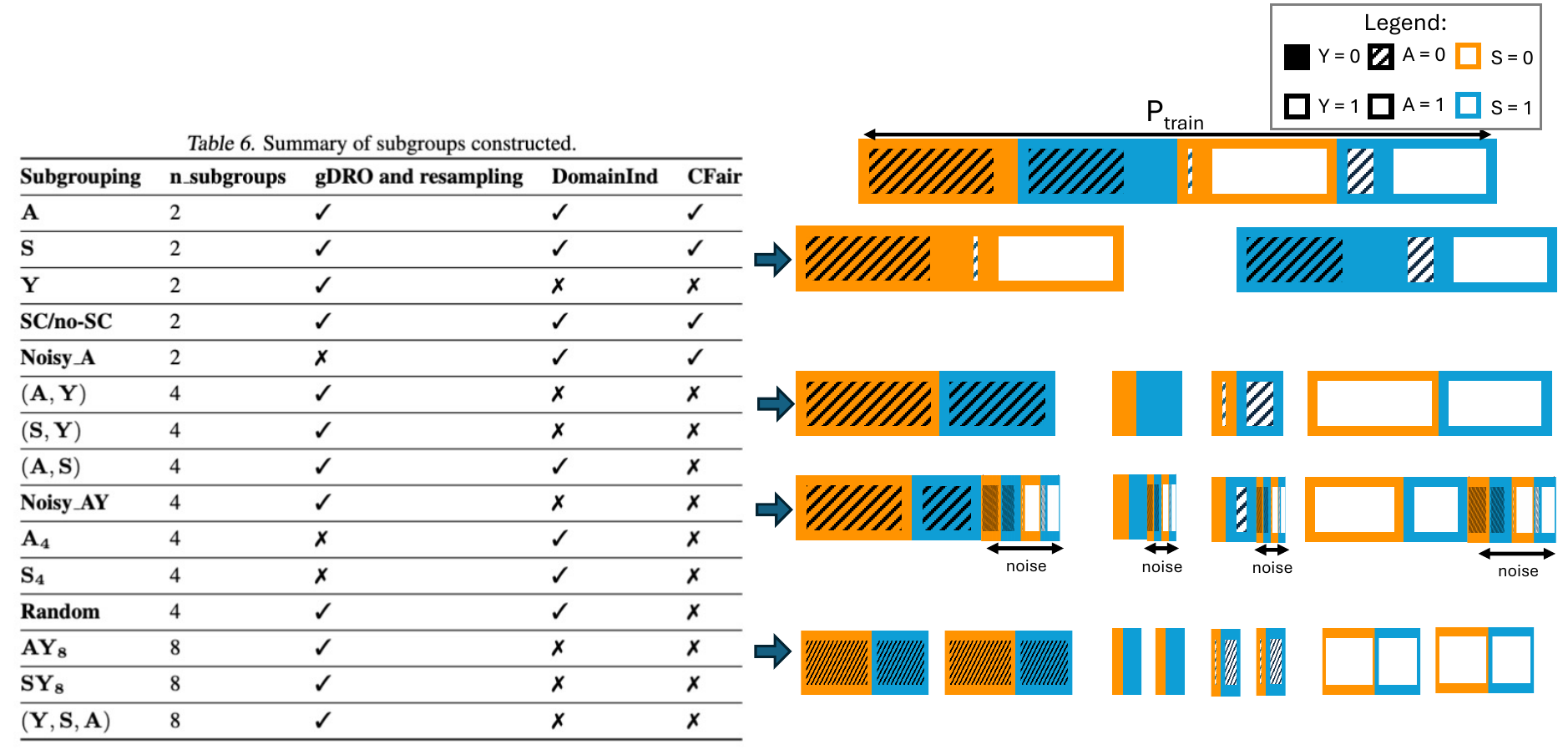}}
    \caption{Subgroup construction for our experiments.}
    \label{fig:subgroup construction diagram}
\end{figure*}
\stepcounter{table}

We show the subgroups used for each method and a visualisation of some example subgroups in Figure \ref{fig:subgroup construction diagram}. For model-based methods (DomainInd and CFair), we do not use $Y$ to construct subgroups because for these methods to work best, each subgroup should contain both positive and negative classes. This is because methods like DomainInd and CFair learn representations for each subgroup separately. DomainInd trains a separate classifier for each subgroup, so it would not make sense to train a separate classification head for positive and negative classes. Similarly, CFair seeks to align subgroup representations, so it would not make sense to align representations of one subgroup containing only positive images to another subgroup containing only negative images, as this would defeat the point of training a discriminative classifier. On the other hand, for reweighting based methods, including the $Y$ in the subgroups helps to balance the final reweighted dataset with respect to class, and therefore improves results, especially in our case where the spurious correlation involves the class $Y$. This explains why we find that the subgroups which work well for DomainInd and CFair (e.g. $A$) are just a merged version of the ones which work well for gDRO and resampling (e.g. $(A,Y)$). To the best of our knowledge, no papers have explicitly discussed this distinction despite its practical importance.

\subsection{Implementation details}
\begin{table}[htb]
\caption{Implementation details for all models.}
\label{tab: implementation details}
\begin{small}
\begin{center}
\resizebox{\columnwidth}{!}{%
\begin{tabular}{@{}lllll@{}}
\toprule
\textbf{Training strategy}  & \textbf{MNIST}           & \textbf{CXP} & \textbf{CelebA} & \textbf{Civil\textunderscore comments}                \\ \midrule
\textbf{Backbone}         & 2-layer CNN                                     & DenseNet121 \cite{huang2018denselyconnectedconvolutionalnetworks} & ResNet50 \cite{he2016deep} & BERTClassifier (uncased) \cite{bertclassifier}                                    \\ \midrule
\textbf{Pre-training}         & None                                    & ImageNet \cite{dengimagenet2009}  & ImageNet \cite{dengimagenet2009} & Bookcorpus, Wikipedia (English)                                     \\ \midrule
\textbf{Batch size}         & 128                                     & 256 & 256 & 32                                        \\ \midrule
\textbf{Image size}         & 3x28x28                                 & 3x299x299 & 3x256x256 & NA                                 \\ \midrule
\textbf{Augmentation}       & Flip, rotation, Gaussian blur    & Flip, rotation, color jitter, affine transformation, crop & Flip, rotation, color jitter, affine transformation, crop & None \\ \midrule
\textbf{Optimiser}          & Adam                                    & Adam  & Adam & AdamW                                     \\ \midrule
\textbf{Loss}               & Binary cross-entropy                    & Binary cross-entropy  & Binary cross-entropy & Binary cross-entropy                       \\ \midrule
\textbf{Learning rate}      & 0.001                                   & 0.0005   & 0.001 & 0.00005                                    \\ \midrule
\textbf{Learning scheduler} & StepLR ($\gamma$ = 0.1 and $\mu$ = 10) & StepLR ($\gamma$ = 0.1 and $\mu$ = 10)   & StepLR ($\gamma$ = 0.1 and $\mu$ = 10) & StepLR ($\gamma$ = 0.1 and $\mu$ = 10)   \\ \midrule
\textbf{Weight decay}       & 0.0001                                  & 0.0001     & 0.0001 & 0.0001                                  \\ \midrule
\textbf{Max epochs}         & 50                                      & 100 (early stopping after 10) & 10 (early stopping after 5) & 10 (early stopping after 5)          \\ \bottomrule
\end{tabular}%
}
\end{center}
\end{small}
\end{table}
We conducted hyperparameter tuning on the baseline ERM model within the ranges below, and selected the model with the highest validation AUC. The choice of backbones was based on their strong performance in previous similar work \cite{irvinCheXpertLargeChest2019,NEURIPS2024_abbbb25c,izmailovfeaturelearning,kirichenko2023layerretrainingsufficientrobustness,idrissiSimpleDataBalancing2022}.

\begin{itemize}
    \item \textbf{Backbones for vision models}: ResNet18, ResNet50, DenseNet121 (not for MNIST images)
    \item \textbf{Batch size}: {32, 64, 128, 256, 512}
    \item \textbf{Learning rate}: [1e-5:1e-3]
    \item \textbf{Weight decay}: [1e-5:1e-4] 
\end{itemize}

We also specify additional hyperparameters for the mitigation methods: a step size of 0.01 and a size adjustment factor of 1 was used for gDRO, and a $\mu$ coefficient of 0.1 was used for the adversarial loss of CFair, following the MEDFAIR implementation \cite{zongMEDFAIRBENCHMARKINGFAIRNESS2023}.

\clearpage

\section{Results for more mitigation methods: DomainInd and CFair}
\label{subsec: results more mitigation}
\begin{figure*}[h]
    \centering
{\includegraphics[height=7cm]{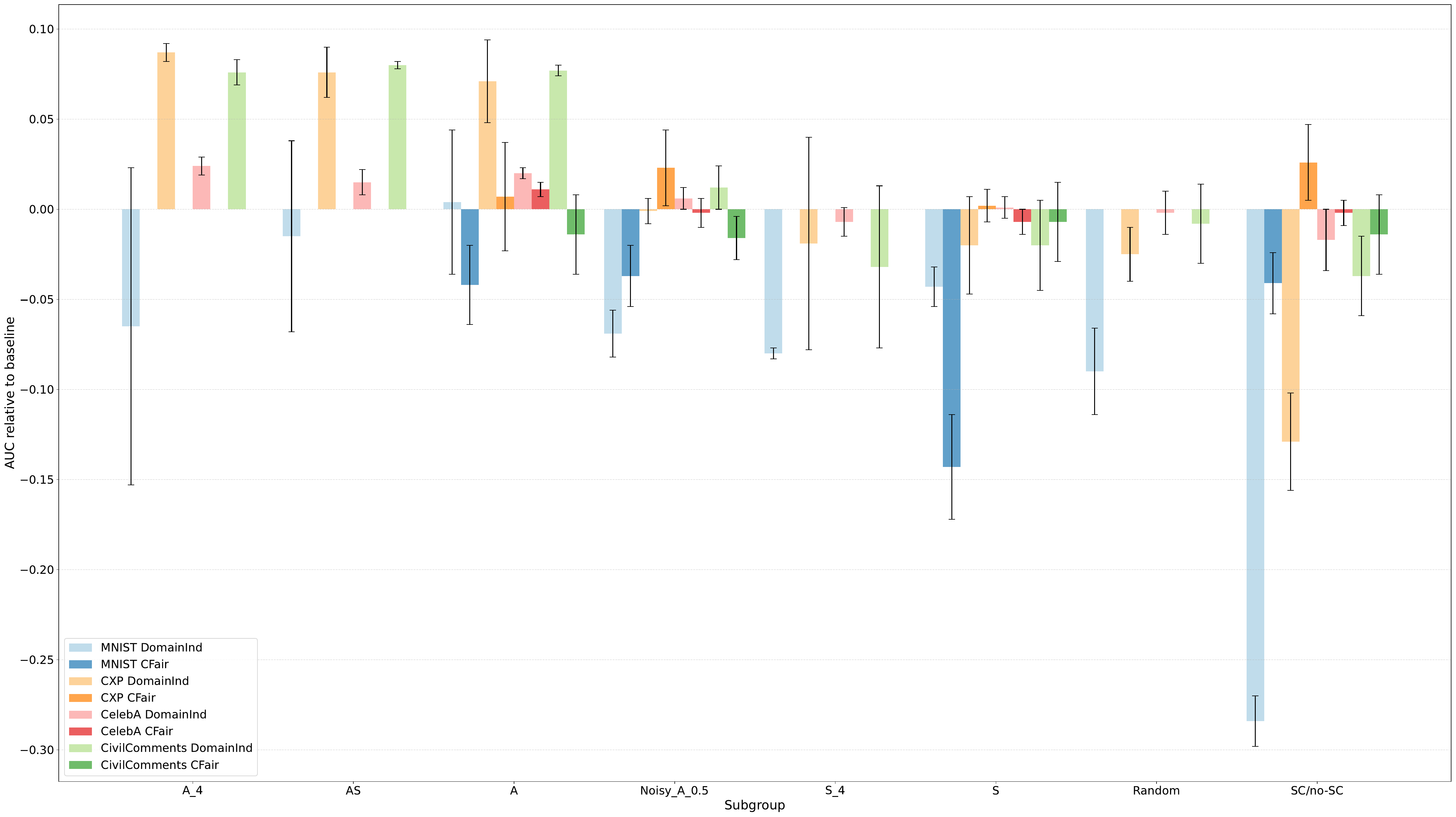}}
    \caption{Relative performance on $\mathcal{P}_{unbiased}$ for different groupings in DomainInd and CFair across all four datasets. Similar trends to gDRO and resampling can be seen, where subgroupings constructed around $A$ generally improve performance as they prevent the SC from being learnt, while other subgroups are generally detrimental. Error bars indicate the standard deviation across 3 random seeds.}
    \label{fig:mnist cxp domainind cfair auc}
\end{figure*}

\begin{figure*}[h]
    \centering
{\includegraphics[height=7cm]{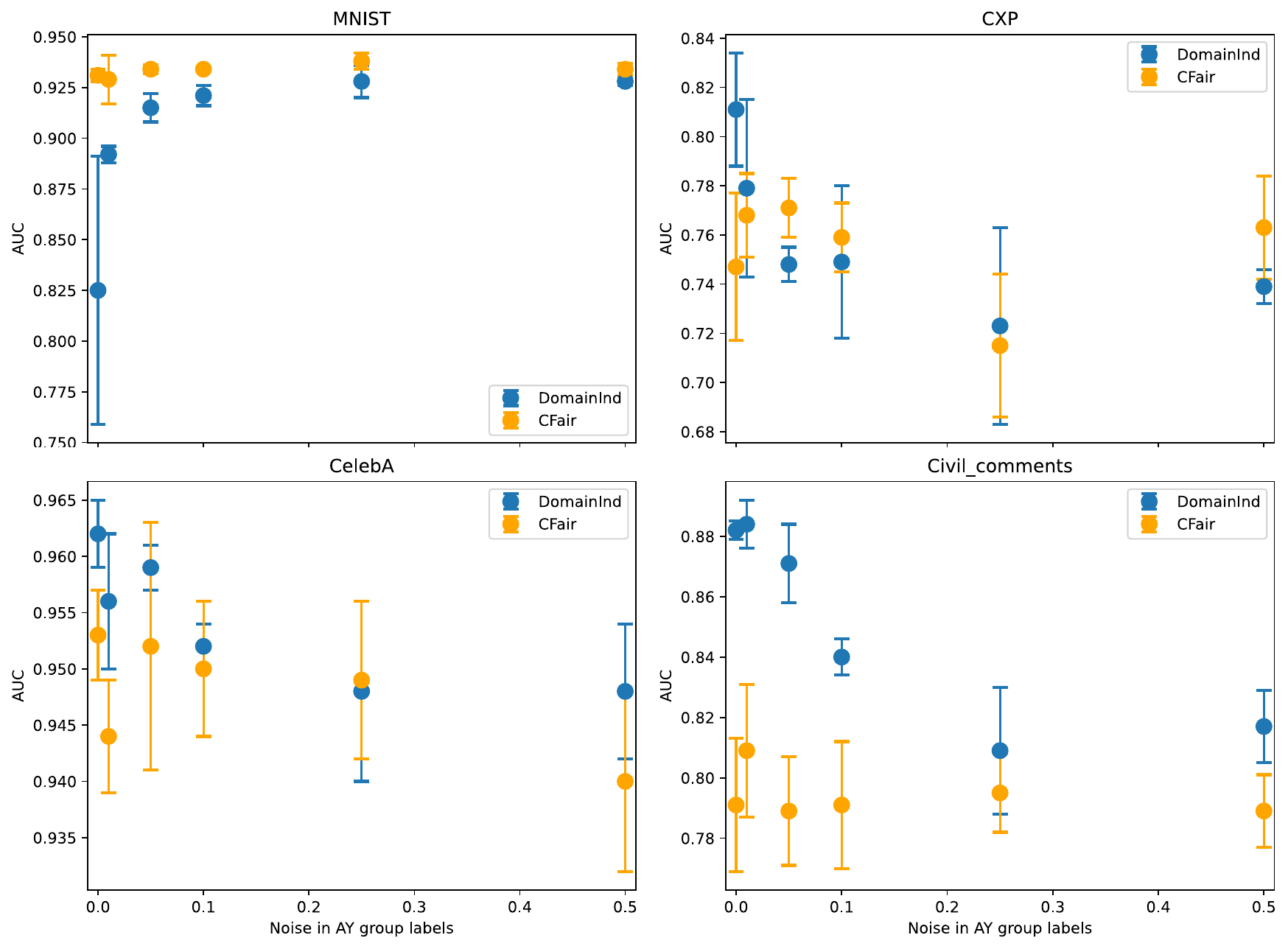}}
    \caption{Effect of noise in $A$ subgroup labels on AUC for DomainInd and CFair. Each dot represents mean performance on the unbiased test set for a specific grouping, with error bars indicating the standard deviation across 3 random seeds.}
    \label{fig:mnist cxp noise auc domainind resampling}
\end{figure*}

Overall, CFair and DomainInd show less improvement on the unbiased test set than reweighting based-methods (Figure \ref{fig:mnist cxp domainind cfair auc}). Despite this, we still observe similar trends as for the reweighting based methods, such as $S$-related groups being clearly detrimental to performance. Learning independent models for $A$ groups also boosts performance for DomainInd in CXP, CelebA, and Civil\_Comments, and while it does not significantly change performance relative to the baseline in MNIST, it is still higher than for any other grouping. Moreover, as shown in Figure \ref{fig:mnist cxp noise auc domainind resampling}, DomainInd appears sensitive to even low levels of noise, while CFair's performance is less degraded by noise in the $A$ subgroup labels (except for in MNIST, where both methods are ineffective, as performance stays close to the baseline for all subgroupings).

\clearpage

\section{Subgroup discovery methods}
\label{subsec: jtt}

We implement Just Train Twice (JTT) as proposed by  \citet{liuJustTrainTwice2021}. JTT consists of a two-stage process, where first a standard ERM model is trained for several epochs, and then a second model that upweights the training examples that the first model misclassified is trained. Although it does not require subgroup labels for training, to select a final model to use (e.g. based on the JTT-specific hyperparameters), subgroup labels do need to be used.

\begin{figure*}[ht]
    \centering
{\includegraphics[height=8cm]{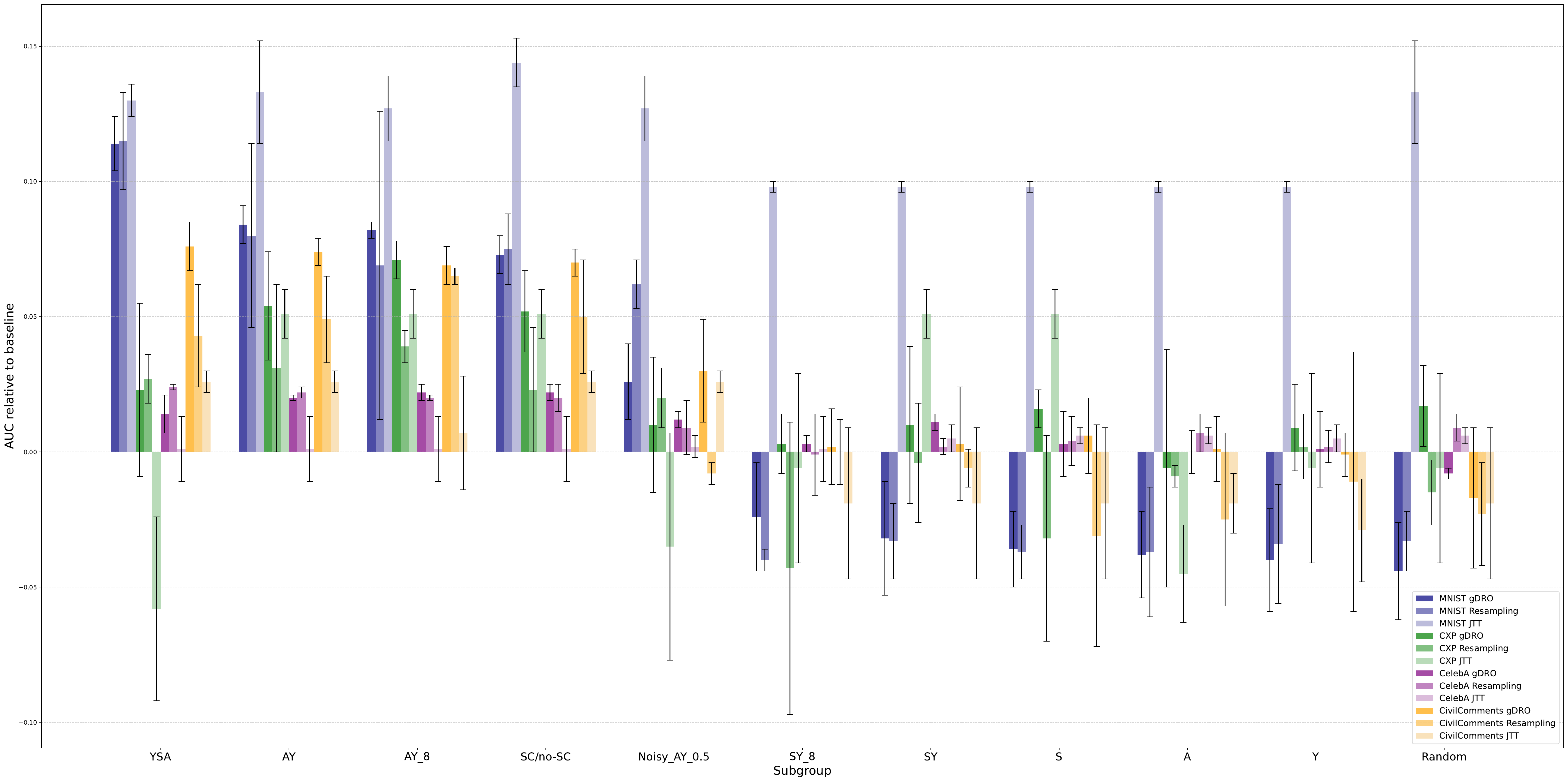}}
    \caption{Performance on the unbiased test set in gDRO and resampling and JTT is highly dependent on the subgroups used. Bars represent overall change in AUC relative to the ERM baseline, with error bars indicating the standard deviation across 3 random seeds.}
    \label{fig: full graph with jtt}
\end{figure*}

As shown in Table \ref{tab: jtt} and Figure \ref{fig: full graph with jtt}, we find that with validation subgroup labels to guide model and hyperparameter selection JTT performs mostly on par with our other methods, however, performance is again highly dependent on the choice of subgroups. When no subgroup annotations are used (i.e. model selection is done by overall validation accuracy), the method does not improve over ERM (except for on MNIST where JTT works remarkably effectively, most likely due to the simplicity of the task). 

\begin{table}[ht]
\begin{small}
\begin{center}
\caption{Just train twice generalisation performance on unbiased test set across the four datasets is highly variable depending on the validation set subgroups used for model/hyper-parameter selection. We colour the experiments which improve over the baseline (no mitigation) in green and those do not in red. We report mean AUC and standard deviation across three random seeds. }
\label{tab: jtt}
\resizebox{0.8\textwidth}{!}{
\begin{tabular}{l|l|l|l|l}
\toprule
\rowcolor{white} \textbf{Subgroup} & \textbf{MNIST}        & \textbf{CXP}         & \textbf{CelebA}       & \textbf{civil\_comments}  \\ \toprule
Baseline  & 0.792 ± 0.057          & 0.740 ± 0.002          & 0.943 ± 0.002         & 0.805 ± 0.017            \\
\rowcolor{white} SY      & \cellcolor[HTML]{C6EFCE}0.89 ± 0.002  & \cellcolor[HTML]{C6EFCE}0.791 ± 0.009  & \cellcolor[HTML]{C6EFCE}0.947 ± 0.005  & \cellcolor[HTML]{FFCCCC}0.786 ± 0.028  \\ 
\rowcolor{white} AY      & \cellcolor[HTML]{C6EFCE}0.925 ± 0.019  & \cellcolor[HTML]{C6EFCE}0.791 ± 0.009  & \cellcolor[HTML]{C6EFCE}0.943 ± 0.012  & \cellcolor[HTML]{C6EFCE}0.831 ± 0.004  \\ 
\rowcolor{white} A       & \cellcolor[HTML]{C6EFCE}0.89 ± 0.002  & \cellcolor[HTML]{FFCCCC}0.695 ± 0.018  & \cellcolor[HTML]{C6EFCE}0.948 ± 0.003  & \cellcolor[HTML]{FFCCCC}0.786 ± 0.011  \\ 
\rowcolor{white} AY\_8   & \cellcolor[HTML]{C6EFCE}0.919 ± 0.012  & \cellcolor[HTML]{C6EFCE}0.791 ± 0.009  & \cellcolor[HTML]{FFCCCC}0.943 ± 0.012  & \cellcolor[HTML]{C6EFCE}0.812 ± 0.021  \\ 
\rowcolor{white} S       & \cellcolor[HTML]{C6EFCE}0.89 ± 0.002  & \cellcolor[HTML]{C6EFCE}0.791 ± 0.009  & \cellcolor[HTML]{C6EFCE}0.948 ± 0.003  & \cellcolor[HTML]{FFCCCC}0.786 ± 0.028  \\ 
\rowcolor{white} SY\_8   & \cellcolor[HTML]{C6EFCE}0.89 ± 0.002  & \cellcolor[HTML]{FFCCCC}0.734 ± 0.035  & \cellcolor[HTML]{FFCCCC}0.943 ± 0.012  & \cellcolor[HTML]{FFCCCC}0.786 ± 0.028  \\ 
\rowcolor{white} Y       & \cellcolor[HTML]{C6EFCE}0.89 ± 0.002  & \cellcolor[HTML]{FFCCCC}0.734 ± 0.035  & \cellcolor[HTML]{C6EFCE}0.947 ± 0.005  & \cellcolor[HTML]{FFCCCC}0.776 ± 0.019  \\ 
\rowcolor{white} Noisy\_AY\_0.5 & \cellcolor[HTML]{C6EFCE}0.919 ± 0.012  & \cellcolor[HTML]{FFCCCC}0.705 ± 0.042  & \cellcolor[HTML]{C6EFCE}0.944 ± 0.004  & \cellcolor[HTML]{C6EFCE}0.831 ± 0.004  \\ 
\rowcolor{white} Random  & \cellcolor[HTML]{C6EFCE}0.925 ± 0.019  & \cellcolor[HTML]{FFCCCC}0.734 ± 0.035  & \cellcolor[HTML]{C6EFCE}0.948 ± 0.003  & \cellcolor[HTML]{FFCCCC}0.786 ± 0.028  \\ 
\rowcolor{white} SC/no-SC & \cellcolor[HTML]{C6EFCE}0.936 ± 0.009  & \cellcolor[HTML]{C6EFCE}0.791 ± 0.009  & \cellcolor[HTML]{FFCCCC}0.943 ± 0.012  & \cellcolor[HTML]{C6EFCE}0.831 ± 0.004  \\ 
\rowcolor{white} YSA     & \cellcolor[HTML]{C6EFCE}0.922 ± 0.006  & \cellcolor[HTML]{FFCCCC}0.682 ± 0.034  & \cellcolor[HTML]{FFCCCC}0.943 ± 0.012  & \cellcolor[HTML]{C6EFCE}0.831 ± 0.004  \\
\rowcolor{white} No val subgroups & \cellcolor[HTML]{C6EFCE}0.925 ± 0.019  & \cellcolor[HTML]{FFCCCC}0.734 ± 0.035  & \cellcolor[HTML]{C6EFCE}0.948 ± 0.003  & \cellcolor[HTML]{FFCCCC}0.786 ± 0.028  \\ \bottomrule
\end{tabular}
}
\end{center}
\end{small}
\end{table}

\clearpage

\section{Supplementary results for resampling and gDRO}
\label{subsec: supplementary results resampling and gdro}

\begin{figure}[H]
\centering
{\includegraphics[height=21.5cm]{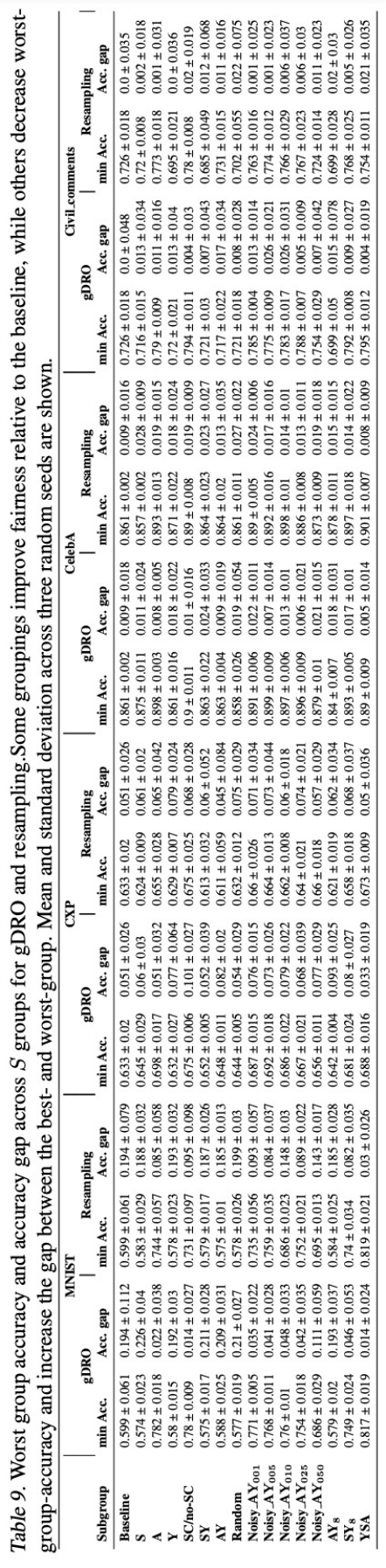}}
    \label{tab:s disparities full table}
\end{figure}

\clearpage

\section{KL divergence between $\mathcal{P}_{\text{train}}^{w}$ and $\mathcal{P}_{\text{unbiased}}$}
\label{sec: kl div calculations}
\subsection{Method overview}

Our objective is to measure the minimum KL divergence which can be achieved to $\mathcal{P}_{unbiased}$ by partitioning $\mathcal{P}_{\text{train}}$ into subgroups and re-weighting these subgroups.

We define each distribution as probability vectors in \( \mathbb{R}^8 \) with each element corresponding to the probability of sampling one $(Y,S,A)$ subgroup. The unbiased distribution is defined to be uniform, i.e., $\mathcal{P}_{\text{unbiased}} = \left[\frac{1}{8}, \frac{1}{8}, \ldots, \frac{1}{8} \right]$, while $\mathcal{P}_{\text{train}}=[
\frac{0.95}{4},\ 
\frac{0.05}{4},\ 
\frac{0.8}{4},\ 
\frac{0.2}{4},\ 
\frac{0.05}{4},\ 
\frac{0.95}{4},\ 
\frac{0.2}{4},\ 
\frac{0.8}{4}
]$.

Initially, $\mathrm{KL}(\mathcal{P}_{\text{train}} \parallel \mathcal{P}_{\text{unbiased}}) 
 \approx 0.527 $. Our aim is to see whether different subgroupings can reduce this divergence.

Let $\mathcal{G} = \{G_1, \dots, G_k\}$ be a partition of the 8 atomic subgroups into $k$ disjoint groups. For a set of weights $w = [w_1, \dots, w_k] \in \Delta^k$ over these groups, we define a new weighted distribution $\mathcal{P}_{\text{train}}^{w} \in \mathbb{R}^8$ as follows:

$$
\mathcal{P}_{\text{train}}^{w}[j] = w_i \cdot \frac{\mathcal{P}_{\text{train}}[j]}{\sum_{l \in G_i} \mathcal{P}_{\text{train}}[l]} \quad \text{for } j \in G_i.
$$

Let the atomic subgroup indices correspond to $(Y,S,A)$ combinations in order $[0,1,2,3,4,5,6,7]$. For the subgroups we constructed, we therefore have:

\begin{itemize}
\item $Y$: $\{\{0,1,2,3\},\{4,5,6,7\}\}$
\item $A$: $\{\{0,2,4,6\},\{1,3,5,7\}\}$
\item $S$: $\{\{0,1,4,5\},\{2,3,6,7\}\}$
\item $(A,Y)$: $\{\{0,2\}, \{1,3\}, \{4,6\}, \{5,7\}\}$
\item $(S,Y)$: $\{\{0,1\}, \{2,3\}, \{4,5\}, \{6,7\}\}$
\item $(Y,S,A)$: $\{0\}, \{1\}, \{2\}, \{3\}, \{4\}, \{5\}, \{6\}, \{7\}$
\item (SC,no-SC): $\{\{0,2,5,7\}, \{1,3,4,6\}\}$
\item Random: $\{\{0,1,2,3,4,5,6,7\}\}$
\item $AY_8$: $\{\{0,2\}, \{0,2\}, \{1,3\}, \{1,3\}, \{4,6\}, \{4,6\}, \{5,7\}, \{5,7\}\}$
\item $SY_8$:  $\{\{0,1\}, \{0,1\}, \{2,3\}, \{2,3\}, \{4,5\}, \{4,5\}, \{6,7\}, \{6,7\}\}$
\end{itemize}

For the $\text{Noisy}_{AY}$ subgroups, we have the same construction as the $(A,Y)$ subgroups, except that $b$ fraction of the $(A,Y)$ subgroup annotations are misannotated following the original $\mathcal{P}_{train}$ distribution, while the remaining $(1-b)$ are consistent with the original $(A,Y)$ subgroups. 

Figure \ref{fig:subgroup construction diagram} provides an illustration of some of these subgroups.

\subsection{Resampling}
For resampling, these weights are uniformly distributed such that $w = [1/k,...,1/k] \in \Delta^k$. We can therefore calculate $\mathcal{P}_{\text{train}}^{w})$ for each grouping keeping the relative proportions of the $(Y,S,A)$ combinations constant within a group. For example, for $Y$, the two groups have probabilities which sum to $\mathcal{P}_{G_1} = \mathcal{P}_{G_2} = \frac{0.95}{4}+\frac{0.05}{4}+\frac{0.8}{4}+\frac{0.2}{4} = \frac{1}{2}$, so the relative proportions within the two subgroups are $[
\frac{\frac{0.95}{4}}{\mathcal{P}_{G_1}},\ 
\frac{\frac{0.05}{4}}{\mathcal{P}_{G_1}},\ 
\frac{\frac{0.8}{4}}{\mathcal{P}_{G_1}},\ 
\frac{\frac{0.2}{4}}{\mathcal{P}_{G_1}},\ 
\frac{\frac{0.05}{4}}{\mathcal{P}_{G_2}},\ 
\frac{\frac{0.95}{4}}{\mathcal{P}_{G_2}},\ 
\frac{\frac{0.2}{4}}{\mathcal{P}_{G_2}},\ 
\frac{\frac{0.8}{4}}{\mathcal{P}_{G_2}},
]$. By multiplying these probabilities by $w = [\frac{1}{2},\frac{1}{2}]$, we get $\mathcal{P}_{\text{train}}^{w} = \mathcal{P}_{\text{train}}$, so $\mathrm{KL}(\mathcal{P}_{\text{train}}^{w} \parallel \mathcal{P}_{\text{unbiased}}) \approx 0.527 $.

We proceed in this way for all subgroups, and obtain the divergences detailed in Table \ref{tab:p_test distance}.

\subsection{gDRO}

For gDRO, the groups are the same but the weights are learned during training. Therefore we determine the weights which could in theory be achieved to give the lowest KL divergence \footnote{We note that these weights may not necessarily be attained in practice by all gDRO models because they are not specifically trained with this objective (although in our setting, minimising KL divergence to the unbiased test set should be a reasonable proxy for minimising worst-group loss). Also, stochasticity in training, optimisation challenges, and inherent difficulties in the task across subgroups may also affect the attainment of this optimum. Despite this, we believe doing this calculation still provides an important indication of the \textit{potential} effectiveness of a chosen subgrouping.}. 
To do this, we reframe the problem as a convex optimisation problem where we minimize the following objective:
\[
\begin{aligned}
\min_{w \in \Delta^k} \quad & \mathrm{KL}\big(\mathcal{P}_{\text{train}}^{w} \parallel \mathcal{P}_{\text{unbiased}}\big) \\
\text{subject to} \quad & w_i > 0, \quad \sum_{i=1}^k w_i = 1.
\end{aligned}
\]
For each subgrouping, we calculate the relative probabilities within a subgroup (as for resampling) and then
use \texttt{scipy.optimize.minimize} and \texttt{scipy.special.rel\_entr} to determine the optimal weight vector subject to the constraints above. 

For example, for $(A,Y)$ subgroups, $\mathcal{P}_{G_1} = \mathcal{P}_{G_4} = \frac{0.95}{4}+\frac{0.8}{4} = \frac{1.75}{4}$ and $\mathcal{P}_{G_2} = \mathcal{P}_{G_3} 
= \frac{0.05}{4}+\frac{0.2}{4} = \frac{0.25}{4}$, so the relative proportions within the two subgroups are $[
\frac{\frac{0.95}{4}}{\mathcal{P}_{G_1}},\ 
\frac{\frac{0.05}{4}}{\mathcal{P}_{G_2}},\ 
\frac{\frac{0.8}{4}}{\mathcal{P}_{G_1}},\ 
\frac{\frac{0.2}{4}}{\mathcal{P}_{G_2}},\ 
\frac{\frac{0.05}{4}}{\mathcal{P}_{G_3}},\ 
\frac{\frac{0.95}{4}}{\mathcal{P}_{G_4}},\ 
\frac{\frac{0.2}{4}}{\mathcal{P}_{G_3}},\ 
\frac{\frac{0.8}{4}}{\mathcal{P}_{G_4}},
]$. Minimisation gives $w = [\frac{1}{4},\frac{1}{4},\frac{1}{4},\frac{1}{4}]$, yielding $\mathcal{P}_{\text{train}}^{w} = [0.136,0.050,0.114,0.200,0.050,0.136,0.200,0.114]$, and $\mathrm{KL}(\mathcal{P}_{\text{train}}^{w} \parallel \mathcal{P}_{\text{unbiased}}) \approx 0.113 $.
Complete results for all groupings are presented in Table \ref{tab:p_test distance}. We often find that these weights correspond to those used for resampling.

\clearpage
\section{Correlation between KL divergence and unbiased generalisation}
\label{subsec: correlation kl div generalisation}

\begin{figure}[ht]
    \centering
{\includegraphics[height=16cm]{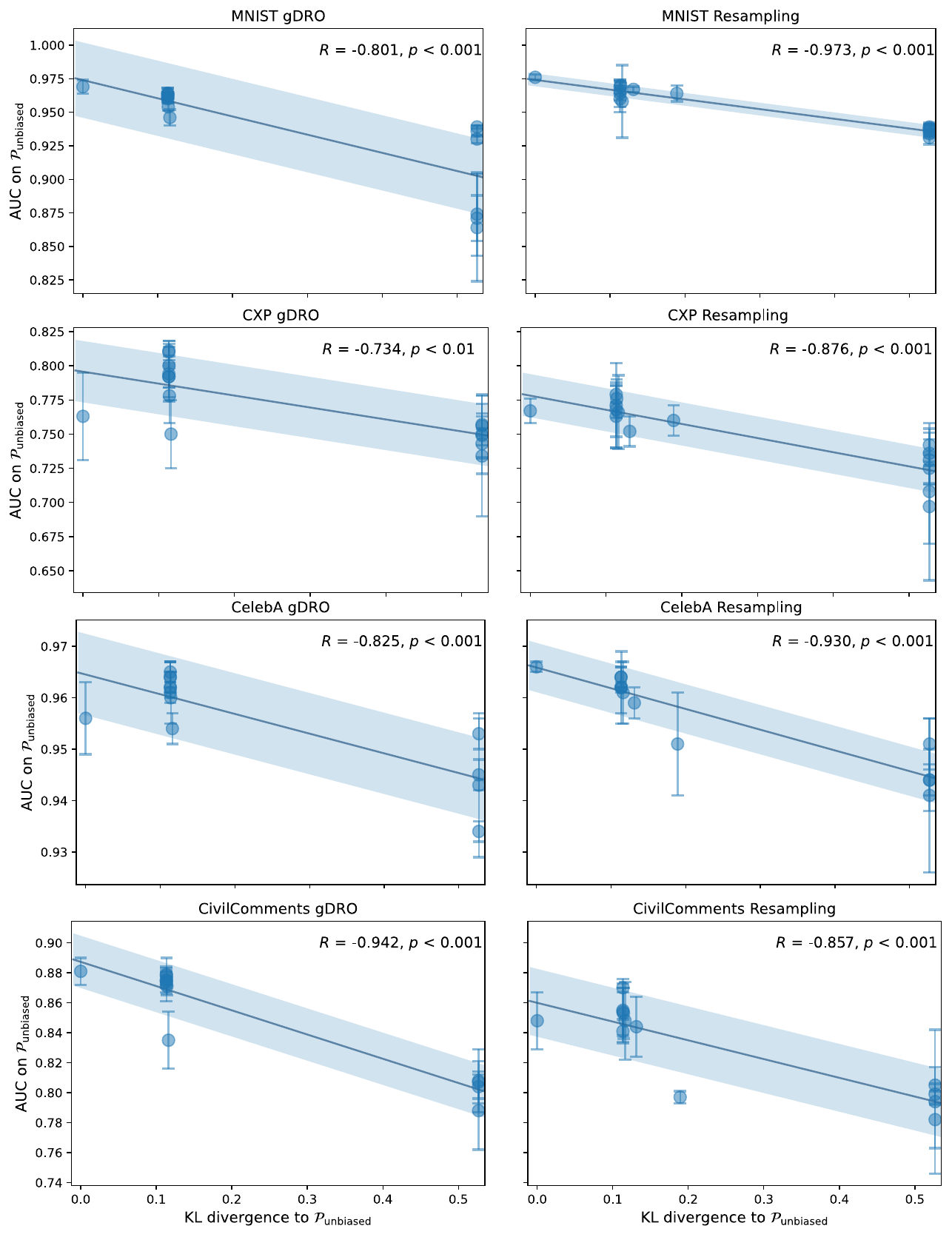}}
    \caption{Test AUC is highly correlated with the minimum achievable KL divergence between $\mathcal{P}_{\text{train}}^{w}$ and $\mathcal{P}_{\text{unbiased}}$ across all four datasets for gDRO and resampling in CXP. Each dot represents mean performance on the unbiased test set for a specific grouping, with error bars indicating the standard deviation across 3 random seeds.}
    \label{fig:kl div auc all datasets}
\end{figure}

\clearpage
\section{Various ablations}
\label{subsec: ablations}
\subsection{MNIST results with a weaker spurious correlation}
\begin{figure*}[htb!]
    \centering
{\includegraphics[height=5cm]{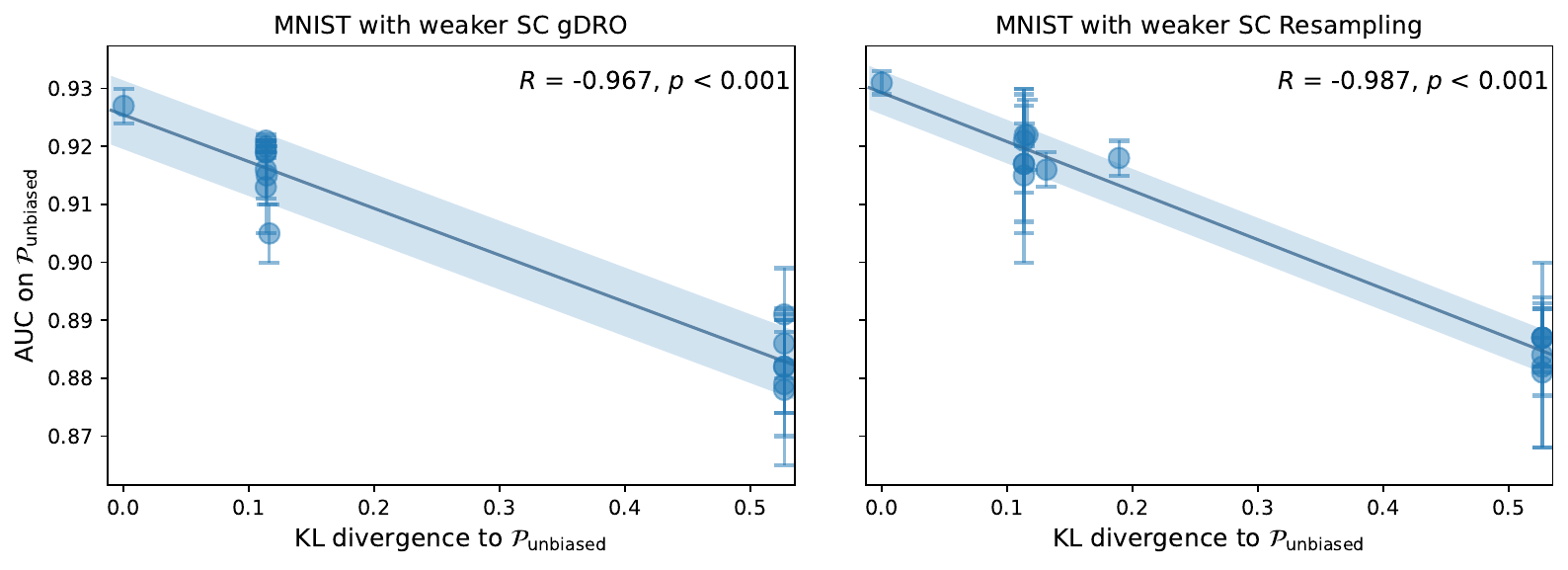}}
    \caption{Relationship between AUC and the minimum achievable distance to the unbiased test distribution ($P_{unbiased}$) for gDRO and resampling in MNIST with a weaker spurious correlation. Each dot represents mean performance on the unbiased test set for a specific grouping, with error bars indicating the standard deviation across 3 random seeds.}
    \label{fig:mnist 85_70 p_test distance auc}
\end{figure*}

To verify that our results still hold in settings where the spurious correlation is weaker, we re-generate the MNIST dataset in the exact same way, except that $P(Y=0,A=0\mid S=0) = P(Y=1,A=1\mid S=0) = 0.85$ and $P(Y=0,A=0\mid S=0) = P(Y=1,A=1\mid S=0) = 0.70$, such that overall there are 77.5\% spuriously correlated samples, instead of 87.5\%. We repeat the same experiments and find that, while all results are higher overall, the same trends still appear. Notably, we identify a significant correlation between the minimum achievable KL divergence to $\mathcal{P}_{unbiased}$ and the overall performance on $\mathcal{P}_{unbiased}$, as shown in Figure G\ref{fig:mnist 85_70 p_test distance auc}. This suggests that subgroup choice is an important factor in less extreme settings of bias as well. 

\subsection{MNIST experiments with a smaller dataset}
\begin{figure*}[htb!]
    \centering
{\includegraphics[height=7.5cm]{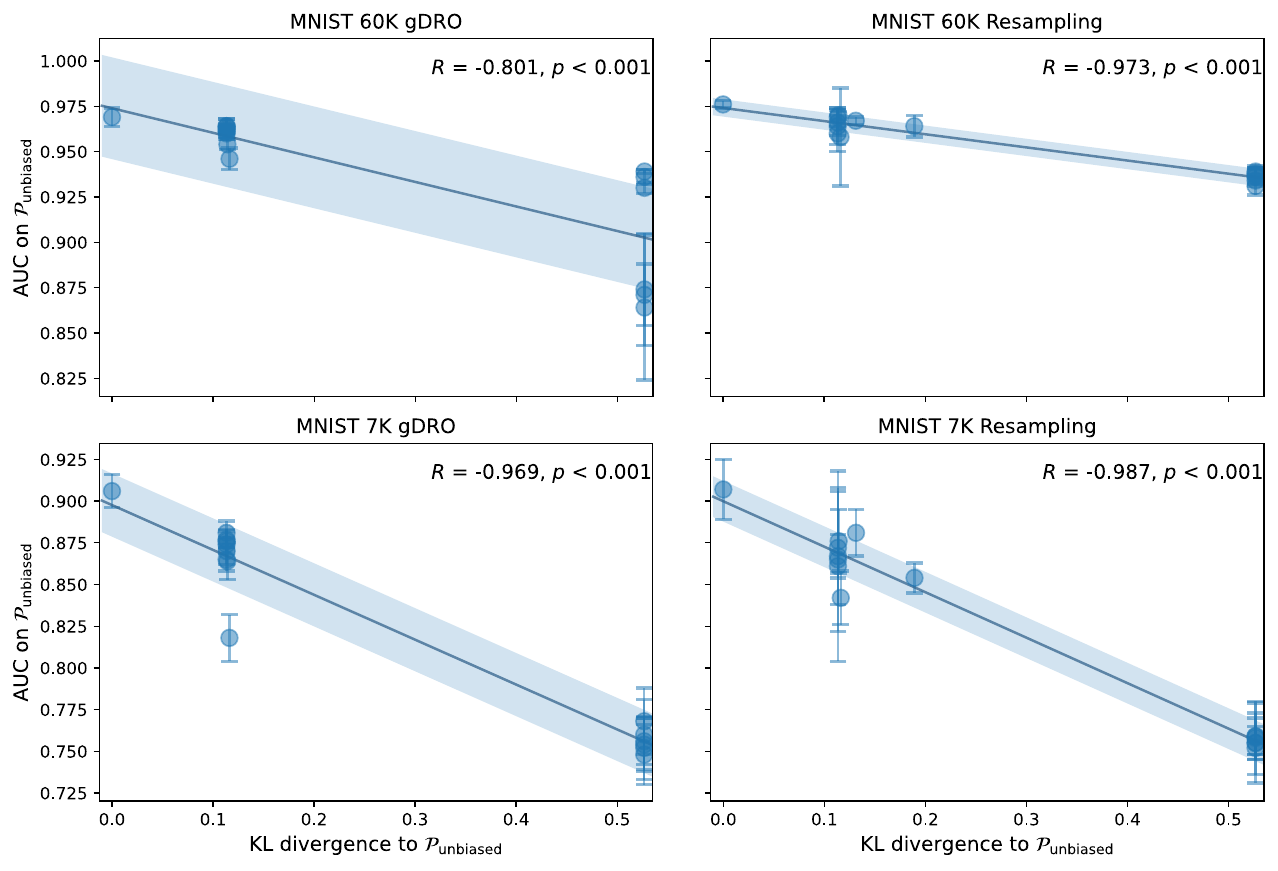}}
    \caption{Relationship between AUC and the minimum achievable KL divergence to the unbiased test distribution ($P_{unbiased}$) for gDRO and resampling in MNIST with a downsampled dataset. Trends appear similar across both dataset sizes, suggesting that the results on the other three datasets would hold had we been able to use a larger subgroup-annotated dataset. Each dot represents mean performance on the unbiased test set for a specific grouping, with error bars indicating the standard deviation across 3 random seeds.}
    \label{fig:mnist 7k 60k kl div auc}
\end{figure*}

\onecolumn


\end{document}